\documentclass[]{template}

\usepackage[utf8]{inputenc}             
\usepackage[T1]{fontenc}                
\usepackage{url}                        
\usepackage{booktabs}                   
\usepackage{multirow}
\usepackage{colortbl}
\usepackage{multicol}
\usepackage{amsfonts}                   
\usepackage{nicefrac}                   
\usepackage{microtype}                  
\usepackage[dvipsnames]{xcolor}         

\usepackage{latexsym}

\usepackage{graphicx}
\usepackage{float}
\usepackage{subcaption}
\usepackage{wrapfig}
\usepackage{lipsum}
\usepackage{adjustbox}

\usepackage{bm}

\usepackage{tabularx} 
\usepackage{ragged2e} 
\newcolumntype{L}{>{\RaggedRight\hangafter=1\hangindent=0em}X}

\usepackage{enumitem}

\usepackage{amsmath}
\usepackage{amssymb}
\usepackage{mathtools}
\usepackage{amsthm}

\setboolean{logo}{true}    

\usepackage[linesnumbered,ruled,vlined]{algorithm2e}

\hypersetup{
    colorlinks=true,
    linkcolor=red,
    citecolor=Cerulean,
    filecolor=magenta,      
    urlcolor=magenta,
}

\usepackage[capitalize,noabbrev]{cleveref}
\crefname{section}{§}{§§}
\Crefname{section}{§}{§§}

\usepackage{calligra}
\DeclareMathAlphabet{\mathcalligra}{T1}{calligra}{m}{n}

\usepackage{pifont}

\theoremstyle{plain}

\theoremstyle{definition}

\theoremstyle{remark}

\renewcommand{\paragraph}[1]{\vspace{1mm}\noindent\textbf{#1}}

\DeclareCaptionLabelFormat{cont}{#1~#2\alph{ContinuedFloat}}
\usepackage[most]{tcolorbox}
\tcbset{
  promptbox/.style={
    top=10pt,
    colback=lightgray!20,
    colframe=Black,
    colbacktitle=NavyBlue,
    enhanced,
    center,
    attach boxed title to top center={yshift=-0.1in,xshift=0.0in},
    boxed title style={boxrule=0pt,colframe=white,},
  }
}
\newtcolorbox{promptbox}[2][]{promptbox, title=#2,#1}
\tcbset{
  takeawaybox/.style={
    top=10pt,
    colback=lightgray!20,
    colframe=Black,
    colbacktitle=BurntOrange,
    enhanced,
    center,
    attach boxed title to top center={yshift=-0.1in,xshift=0.0in},
    boxed title style={boxrule=0pt,colframe=white,},
  }
}
\newtcolorbox{takeawaybox}[2][]{takeawaybox, title=#2,#1}
\tcbset{
  observationbox/.style={
    top=10pt,
    colback=lightgray!20,
    colframe=Black,
    colbacktitle=YellowGreen,
    enhanced,
    center,
    attach boxed title to top center={yshift=-0.1in,xshift=0.0in},
    boxed title style={boxrule=0pt,colframe=white,},
  }
}
\newtcolorbox{observationbox}[2][]{observationbox, title=#2,#1}

\usepackage{xspace}

\newcommand\blfootnote[1]{%
  \begingroup
  \renewcommand\thefootnote{}\footnote{#1}%
  \addtocounter{footnote}{-1}%
  \endgroup
}

\usepackage{CJK}

\title{Intern-S2-Mobius: Foundation Model with Decoupled Knowledge and Reasoning}

\author{Intern-S2-Mobius Team, Shanghai AI Laboratory}

\begin{abstract}

  We introduce \textbf{\textit{Mobius-v0}}, an architecture that comprises a globally shared Memory (FFN) that stores knowledge vectors and multiple Reasoners (Self-Attn) that iteratively achieve compositional reasoning. Using hidden states as cache and carrier, reasoners repeatedly query memory for required knowledge-vectors, while the knowledge is transmitted back to reasoning operators. Through this knowledge-reasoning-separation architecture, Mobius achieves better knowledge compression and reasoning efficiency. Built upon Mobius-v0 architecture: 1) Our 7B model trained-from-scratch achieves similar downstream score as a 7B Transformer baseline with 62.6\% of baseline's training data. 2) Our \textbf{Intern-S2-Mobius}, continually-pretrained from Qwen3.5-35B, achieves similar downstream score while delivering nearly 4× end-to-end inference speedup.

\end{abstract}

\begin{document}

\blfootnote{$*$ Model is available at \url{https://huggingface.co/internlm/Intern-S2-Mobius}}

\maketitle

\section{The development bottleneck of the Foundation Models}

\begin{figure}[h]
    \centering
    \includegraphics[width=\textwidth]{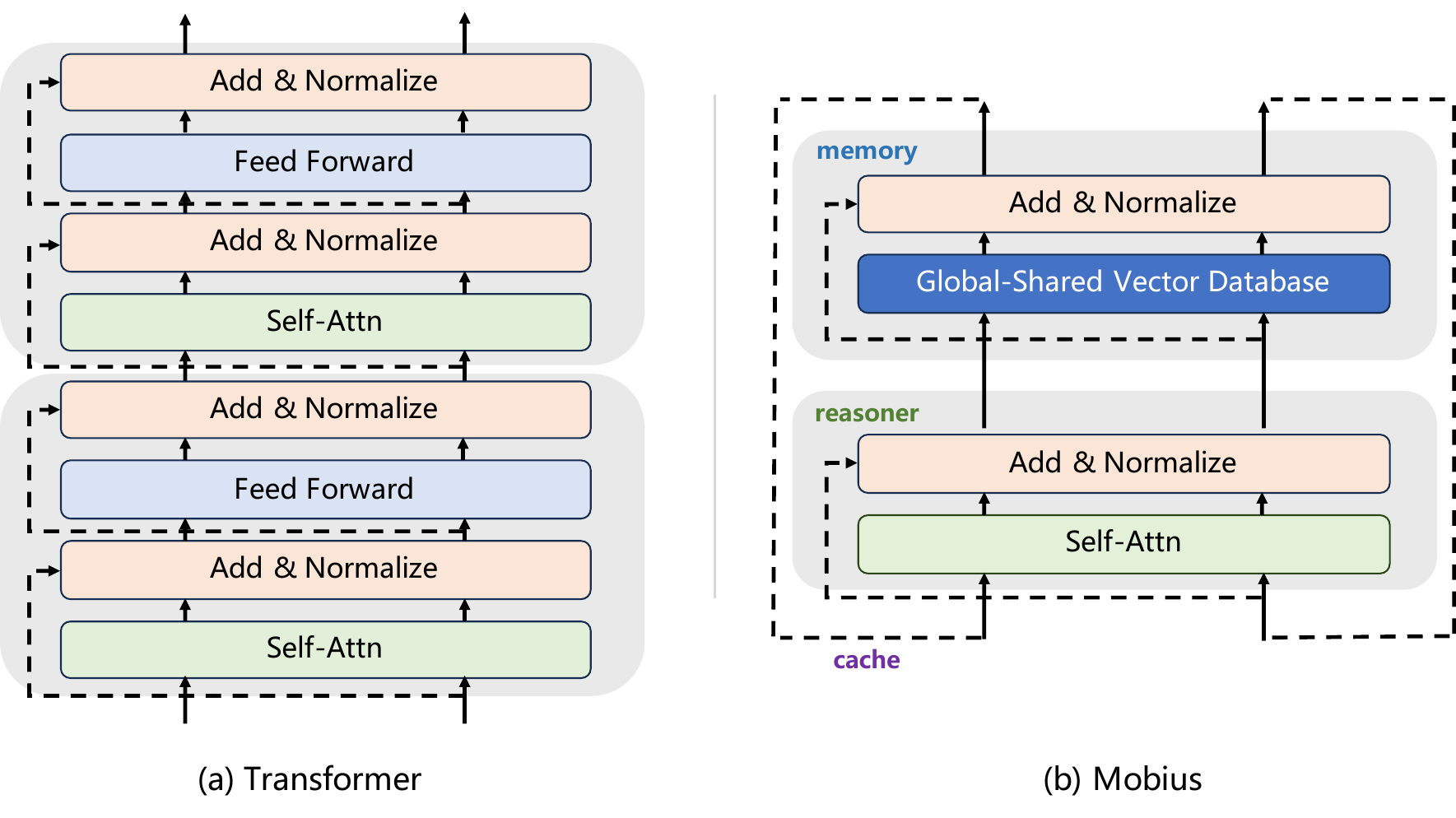} 
    \caption{The Comparison between Transformer and Mobius.}
    \label{fig:architecture}
\end{figure}

The Transformer~\citep{attentionisallyouneed} stands as the most prevalent and powerful architectural paradigm to date, permeating virtually every domain—from language and vision, to video and scientific computing~\citep{bert,vit,video_transformer,zou2026interns1pro}. Along its evolutionary trajectory, two pivotal research directions have emerged. The first preserves the model architecture intact while increasing expenditure elsewhere to enhance capability: scaling model parameters, expanding training corpora, and elongating reasoning chains typically endow models with richer knowledge and the capacity to tackle more intricate problems~\citep{scalinglaws,compute-optimal-llm,Wei0SBIXCLZ22,verifystepbystep}. The second reduces training and inference overhead by lowering architectural complexity to improve practicality: motivated by the widely held view that the quadratic complexity of self-attention constitutes a bottleneck for ultra-long contexts, linear attention mechanisms such as SSM and GDN have been proposed~\citep{lineartransformer,performer,mamba,gdn}, and their variants have since been integrated into mainstream contemporary architectures.

\begin{figure}[ht]
    \centering
    \includegraphics[width=0.8\textwidth]{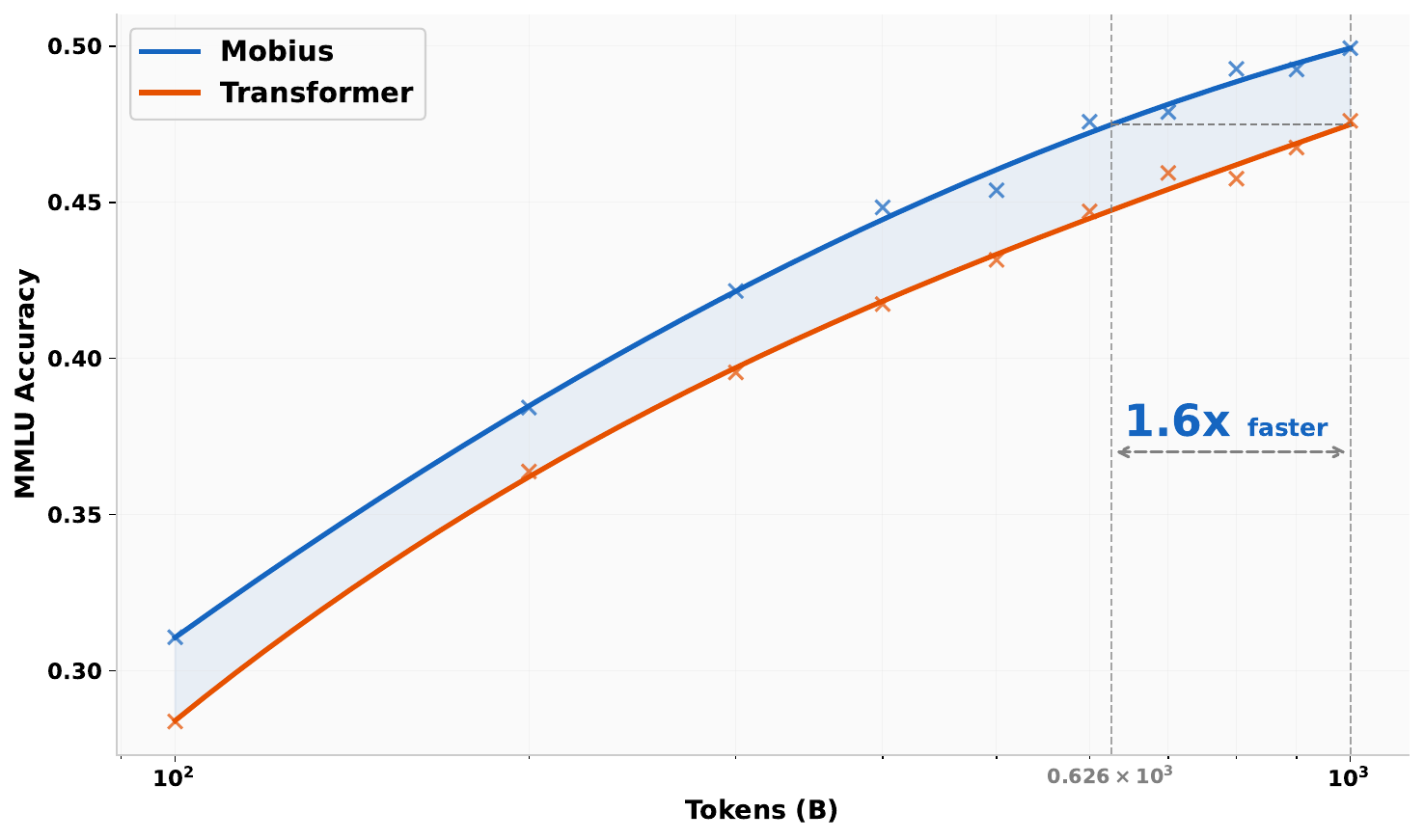} 
    \caption{The MMLU score of Mobius and Transformer pre-trained from scratch.}
    \label{fig:scratch-mmlu}
\end{figure}

\textbf{Yet both approaches are now hitting their limits.} On one hand, scaling data and parameters has yielded increasingly powerful models at high returns on investment, yet the Scaling Law has gradually plateaued and is approaching diminishing marginal returns~\citep{scalinglaws,compute-optimal-llm}. Long chains of thought have served us well in domains such as mathematics, code, and physics, yet models tend to produce verbose, tangential output regardless of problem difficulty—a trait typically regarded in human society as a sign of insufficient intelligence~\citep{simple-test-time-scaling,reasoningmodelseffectivethinking}. On the other hand, reducing the computational complexity of full attention does bring certain efficiency gains, yet it has become increasingly apparent that the efficiency improvements achieved at the cost of sacrificing a substantial portion of model capability have not made models truly affordable enough for commercial deployment~\citep{linearattention-tradeoff}.

Confronted with this bottleneck in AI development, we pursue an alternative path: \textbf{increasing architectural complexity to raise the ceiling of model intelligence, thereby reducing the end-to-end overhead.} To this purpose, we propose the \textbf{\textit{Mobius}}, which decouples the binding between knowledge vectors (FFN) and reasoning operators (Self-Attn), thereby constructing a shared knowledge vector database accessible to all reasoning operators. Although the large scale of this shared vector database renders each inference activation highly sparse, introducing greater memory access pressure and lower per-pass forward efficiency, we find that this implementation achieves nearly 4× end-to-end inference speedup over the Transformer architecture with same parameters while maintaining equivalent reasoning accuracy.

The efficiency gains of Mobius reasoning stem primarily from two sources: a more flexible activation path during inference, and a more dynamic latent-space iteration, which together endow Mobius with a more efficient and concise output pattern. The first advantage arises because, unlike the combination of hierarchical storage and forward residual connections in Transformer, Mobius's shared storage natively introduces \textbf{Backward Residual Connection}. This mechanism means that not only can shallow hidden states access deep-layer knowledge, but deep hidden states can also access shallow-layer knowledge, which enhances the compositional generalization across different layers and thereby accelerates the synthesis of critical information. The second advantage arises because Mobius natively introduces \textbf{Dynamic Latent Reasoning}: whereas Transformer uses tokens as the medium of information transfer and requires traversing all layers to complete the inference of a single token, Mobius can iterate and refine latents against the full knowledge repository within just a few layers, rather than requiring multiple full-layer iterations as in traditional latent reasoning. These latents are not tightly bound to any specific token, and only at deeper layers are multiple tokens decoded synchronously. This approach both increases the density of information transfer and dynamically allocates varying computational costs to different tokens. Ultimately, Mobius completes the same reasoning task with markedly fewer high-quality tokens, achieving substantially higher end-to-end inference efficiency than Transformer.

\definecolor{internblue}{RGB}{220,235,247}

\newcolumntype{C}[1]{>{\centering\arraybackslash}m{#1}}
\newcolumntype{B}[1]{>{\columncolor{internblue}\centering\arraybackslash}m{#1}}

\newcommand{\SubTableTitle}[1]{%
  \multicolumn{3}{l}{\hspace{-2pt}\textbf{#1}} \\
}

\begin{table*}[h]
    \centering
    \caption{Performance comparison across general and scientific benchmarks. The higher score in each row is highlighted in \textbf{bold}.}
    \label{tab:cpt-downstream}
    \small
    \setlength{\tabcolsep}{8pt}
    \begin{adjustbox}{max width=\textwidth}
        \begin{tabular}{l B{3.4cm} C{3.4cm}}
    \SubTableTitle{General Tasks}
    \toprule
    \textbf{Benchmark} & \textbf{Intern-S2-Mobius-35B} & \textbf{Qwen3.5-35B} \\
    \midrule
    MMLU Pro             & \textbf{89.05} & 85.31 \\
    GPQA Diamond         & \textbf{80.81} & 80.24 \\
    IMO Bench            & \textbf{81.25} & 77.50 \\
    AIME 2026            & \textbf{95.31} & 92.08 \\
    HMMT 2026            & \textbf{85.51} & 78.50 \\
    UGD hard             & 73.02 & \textbf{78.02} \\
    AMO                  & \textbf{58.00} & 50.00 \\
    SimpleQA             & \textbf{28.90} & 21.39 \\
    HLE                  & 19.11 & \textbf{22.40} \\
    \bottomrule
    AVG Score        & \textbf{67.88} & 65.05 \\
    \bottomrule \\[-0.5em]
    
    \SubTableTitle{Scientific Tasks}
    \toprule
    \textbf{Benchmark} & \textbf{Intern-S2-Mobius-35B} & \textbf{Qwen3.5-35B} \\
    \midrule
    Biology-Instructions & \textbf{51.40} & 3.77 \\
    Mol-Instructions     & \textbf{45.73} & 21.70 \\
    MolecularIQ          & \textbf{59.29} & 29.13 \\
    \bottomrule
    AVG Score        & \textbf{52.14} & 18.20 \\
    \bottomrule

    \end{tabular}
\end{adjustbox}
\end{table*}

\begin{figure}[h]
    \centering
    \includegraphics[width=\textwidth]{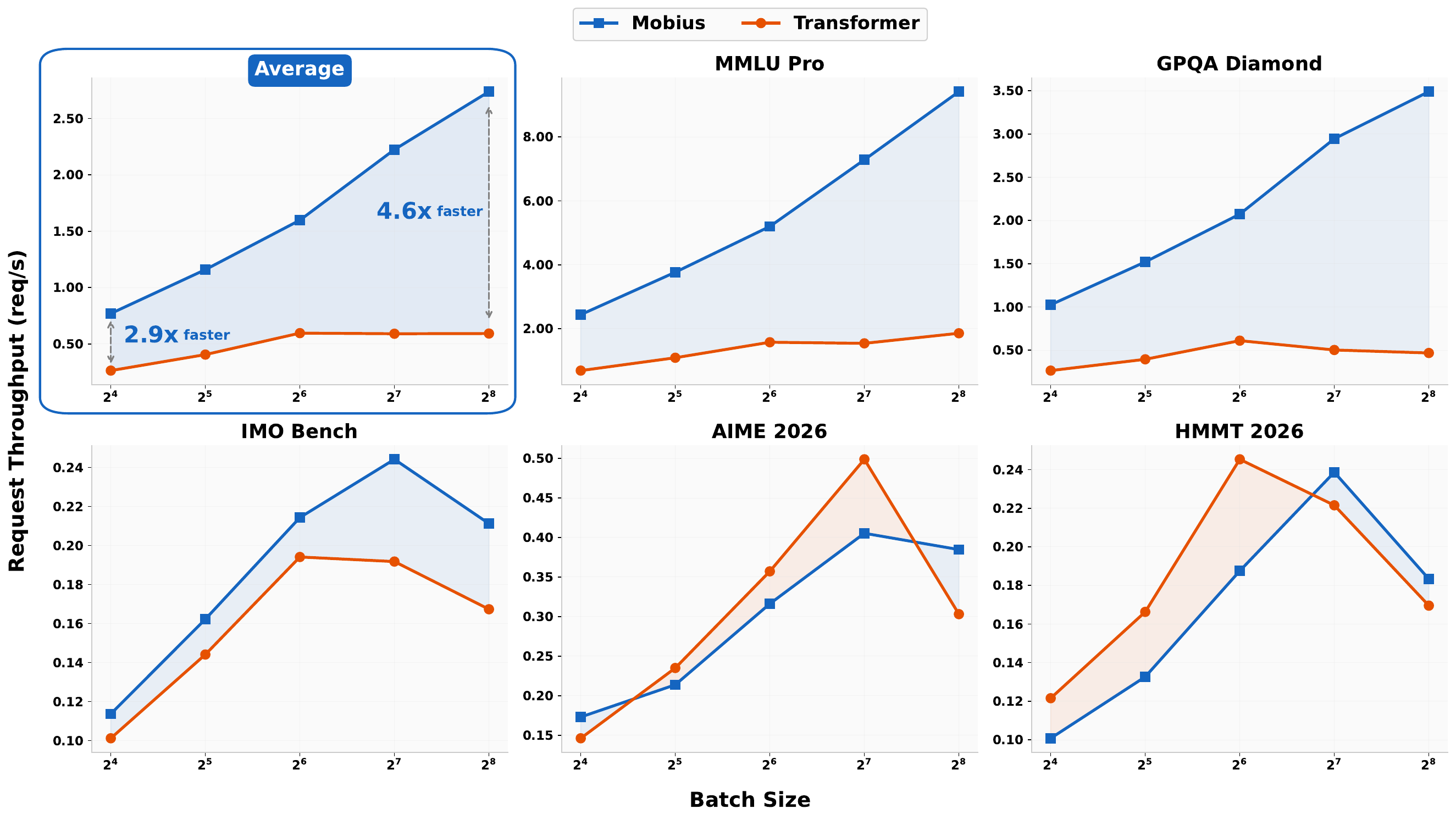} 
    \caption{The inference efficiency of Mobius continual pre-trained from Qwen3.5.}
    \label{fig:cpt-efficiency}
\end{figure}

\begin{figure}[h]
    \centering
    \includegraphics[width=\textwidth]{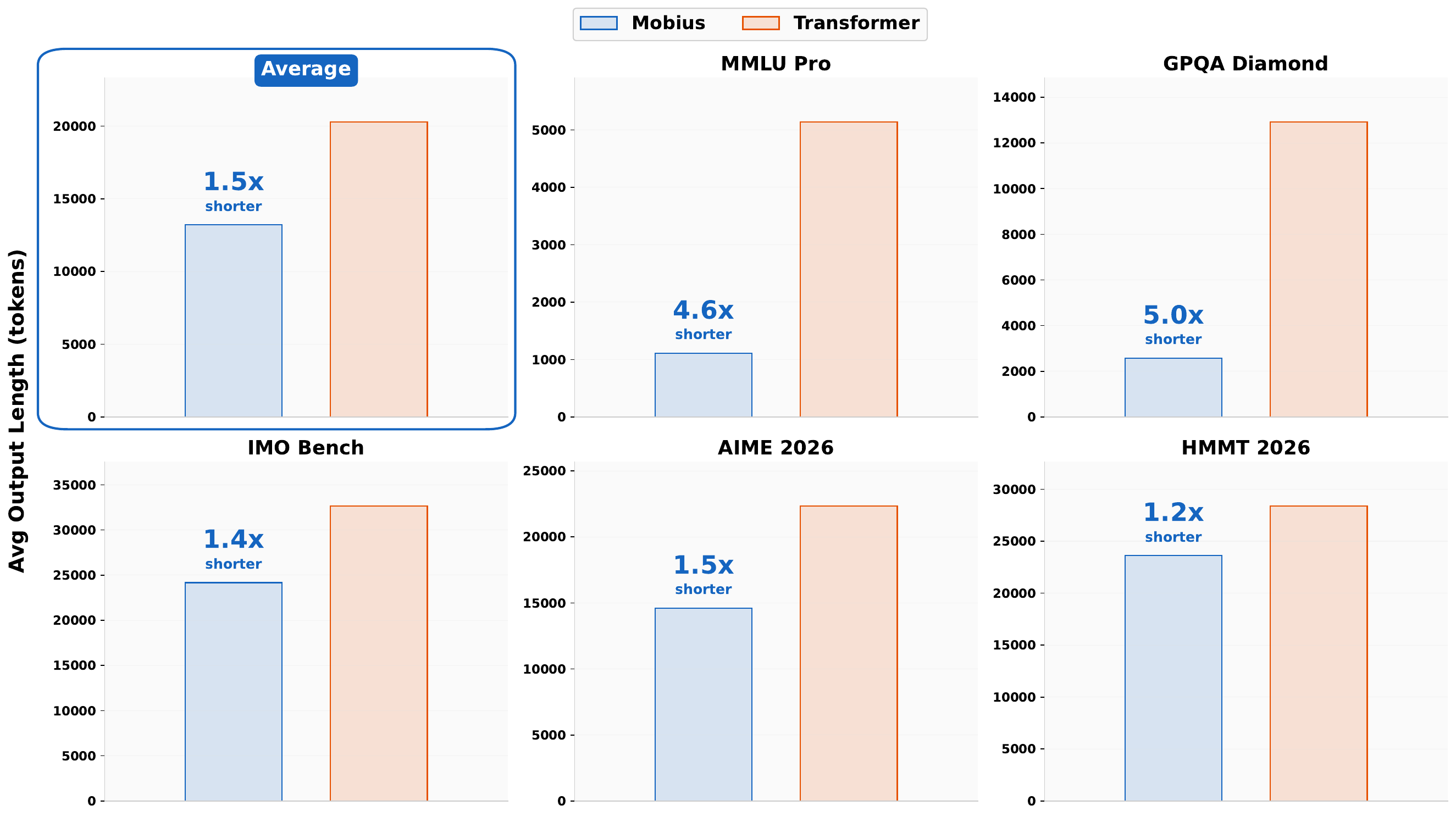} 
    \caption{The average output length of Mobius continual pre-trained from Qwen3.5.}
    \label{fig:cpt-cot}
\end{figure}

\begin{table*}[h]
    \centering
    \caption{Step-aligned comparison between Intern-S2-Mobius-35B and Qwen3.5-35B
    on a linear-algebra multiple-choice question. Both models select the correct
    answer (Option C). Token counts are computed using the Qwen3.5-35B tokenizer. Mobius completes the same reasoning steps with fewer tokens, which mainly benefits from the model's elimination of repeated derivation and checks.}
    \label{tab:linear-algebra-cot-comparison}
    \small
    \setlength{\tabcolsep}{8pt}
    \renewcommand{\arraystretch}{1.12}

    \begin{adjustbox}{max width=\textwidth}
    \begin{tabular}{l B{5.2cm} C{5.2cm}}

    \SubTableTitle{Reasoning Trace Comparison}
    \toprule
    \textbf{Aligned Step} &
    \textbf{Intern-S2-Mobius-35B} &
    \textbf{Qwen3.5-35B} \\
    \midrule

    Task framing &
    \textbf{17 tokens.}

    ``The user wants me to evaluate the truth value of two statements regarding
    linear algebra.'' &
    \textbf{20 tokens.}

    ``The user wants me to evaluate two statements and determine their truth
    values from a list of options.'' \\

    \addlinespace[0.2em]
    Statement 1 &
    \textbf{178 tokens.}

    ``Counter-example: Let $v_1=(1,0)$ and $v_2=(2,0)$. Then
    $2v_1-v_2=0$, so they are linearly dependent. \ldots{} Statement 1 is
    False.'' &
    \textbf{283 tokens.}

    ``Consider the zero vector. Let $v_1=(1,0)$ and $v_2=(0,0)$.
    \ldots{} Consider two parallel vectors. Let $v_1=(1,0)$ and
    $v_2=(2,0)$. \ldots{} Statement 1 is \textbf{False}.'' \\

    \addlinespace[0.2em]
    Statement 2 &
    \textbf{142 tokens.}

    ``A basis is a linearly independent spanning set. Since
    $\{v_1,\dots,v_k\}$ spans $V$ and is linearly independent, it forms a
    basis for $V$. \ldots{} $\dim(V)=k$.'' &
    \textbf{197 tokens.}

    ``The set forms a basis for $V$ if it spans $V$ and is linearly
    independent. \ldots{} Therefore, $\{v_1,\dots,v_k\}$ is a basis for $V$.
    \ldots{} $\dim(V)=k$.'' \\

    \addlinespace[0.2em]
    Option matching &
    \textbf{22 tokens.}

    ``Statement 1 is False, Statement 2 is True. This matches Option C.'' &
    \textbf{122 tokens.}

    ``The combination is False, True. This corresponds to Option C.'' \\

    \addlinespace[0.2em]
    Repeated derivation and checks &
    \textemdash &
    \textbf{1,147 tokens.}

    ``\textbf{Step-by-step derivation:} \ldots{} Counterexample:
    Let $v_1=(1,0)$ and $v_2=(2,0)$. \ldots{}
    \textbf{Final check:} \ldots{} If $k=0$? \ldots{} Works.
    \ldots{} Option C.'' \\

    \addlinespace[0.2em]
    Visible final answer &
    \textbf{157 tokens.}

    ``The vectors $(1,0)$ and $(2,0)$ are linearly dependent because one is a
    scalar multiple of the other. \ldots{} This is the definition of a basis.
    \ldots{}

    ANSWER: C'' &
    \textbf{595 tokens.}

    ``\textbf{Step 1: Evaluate Statement 1} \ldots{} $2v_1-1v_2=(0,0)$.
    \ldots{} Statement 1 is \textbf{False}.

    \textbf{Step 2: Evaluate Statement 2} \ldots{} $\dim(V)$ is equal to $k$.
    \ldots{}

    ANSWER: C'' \\

    \midrule
    \textbf{Total} &
    \textbf{516 tokens} &
    \textbf{2,364 tokens} \\

    \bottomrule
    \end{tabular}
    \end{adjustbox}

    \vspace{0.25em}
\end{table*}

\section{What inspired the design of Mobius?}

In Transformer-based architectures, the Feed-Forward Network is conventionally regarded as responsible for knowledge storage~\citep{key-value-memories,rome}, Hidden States for information transmission, and Self-Attention for compositional reasoning~\citep{attentionisallyouneed,methmatical-framework-for-transforner-circuits}.

Concurrently, owing to the hierarchical structure of Transformer, Self-Attention at each layer primarily processes knowledge inputs received from the preceding layer and induces the FFN at the current layer to produce outputs required by the next layer. Although residual connections establish more flexible connections across layers~\citep{he2016resnet,highway,densenet}, only shallow Hidden States can access deep-layer knowledge, while deep-layer Self-Attention can only process knowledge originating from shallow layers; the model remains incapable of information transfer in the opposite direction.

To construct information transfer in the reverse direction, current models predominantly rely on Chains of Thought~\citep{Wei0SBIXCLZ22,deepseekai2025r1}. After each forward pass, the model generates a new token that serves to activate knowledge in the next inference step. Through iterative token generation, the model continuously extracts valuable knowledge from the FFN until this knowledge suffices to produce critical tokens or even the final answer.

The joint construction of artificial intelligence models using exclusively hierarchical structures, forward residual connections, and token-mediated chains of thought is inefficient. To equip models with sufficient knowledge for answering complex questions, an inordinate amount of computation is expended on producing lengthy, redundant reasoning chains. To mitigate this issue, we have redesigned the model architecture and propose Mobius. 

\subsection{Mobius’ first innate talent —— Backward Residual Connection}

Residual connections have become one of the most critical components in modern deep learning models~\citep{he2016resnet}. Seemingly minor, they bear the crucial responsibility of transmitting inter-layer information during forward propagation and stabilizing gradient magnitudes during backpropagation.~\citep{he2016identity,xiong2020layer} Although numerous variants of residual connections have since emerged, such as Hyper-Connection and Attention-Residual~\citep{zhu2025hyperconnections,mhc,kimi2026attnres}, all mainstream residual to date retain their unidirectional nature. For instance, during forward propagation, residual only convey shallow-layer information to deeper layers.~\citep{densenet} This unidirectionality entails significant drawbacks: if certain critical knowledge fails to be activated during shallow-layer computation, the model may struggle to decode valuable tokens in that reasoning round, and can only resort to generating low-information tokens via chains of thought to proceed to the next reasoning step.

To mitigate this phenomenon, rendering residual connections bidirectional is necessary—that is, enabling deep layers to access shallow-layer knowledge during forward propagation. However, a direct implementation of such backward residual connections is infrastructure-unfriendly, as it may introduce more complex computation graphs and harder-to-parallelize asynchrony. To this end, we opt for an indirect realization of backward residual connections: different layers share a single, oversized knowledge repository, granting every layer the opportunity to access all knowledge within the model. Empirical results ultimately corroborate that this form of backward residual connection contributes meaningfully to improving the model's end-to-end inference efficiency.

\subsection{Mobius’ second innate talent —— Dynamic Latent Reasoning}

Long chains of thought (Long CoT) have become an indispensable component in large language model reasoning~\cite{Wei0SBIXCLZ22,verifystepbystep,deepseekai2025r1}. Empowered by this technique, contemporary large language models can now solve a wide spectrum of complex reasoning problems, spanning mathematics, physics, and code generation. Yet the cost of Long CoT is exorbitant. On one hand, longer reasoning chains entail the generation of more tokens, with generation cost growing non-linearly with chain length. On the other hand, current models frequently adopt a trial-and-error-and-correct approach when tackling complex problems, resulting in substantial token redundancy during extended reasoning. Moreover, such models tend to produce excessively verbose responses even for simple problems. This verbosity stems partly from the residual connections discussed above, and partly from the fact that the minimal unit of our current reasoning is the token—a discrete, low-information-density storage ~\citep{cheng2024compressed}.

Mobius internalizes processes such as deliberation, trial-and-error, and refinement into the optimization of a continuous vector, and enables the model to dynamically allocate computational budget to different tokens. This creates the opportunity for the model to produce fewer superfluous tokens while achieving more efficient and more intelligent reasoning. Mobius can be regarded as an upgraded synthesis of the Looped Transformer and the Diffusion Language Model~\citep{loopedtransformer,li2022diffusionlm,nie2025llada}, employing more efficient latent reasoning and parallel prediction. First, Mobius operates at a higher loop frequency, completing one representation iteration with extremely few layers. Second, during inference, Mobius acquires joint representations of multiple tokens simultaneously through more compact, high-information-density continuous vectors. This, on one hand, alleviates the pressure from parallelizing multiple hidden states, reduces the complexity of KV-cache management during the looping process, and enhances the continuous differentiability of the vector iteration process. Meanwhile, this native latent reasoning endows the model with a more dynamic and unfettered iteration process, without constraining the model to decode a fixed number of tokens at specific iteration steps.

\subsection{One Stone Two Birds —— Disentangling Knowledge Vectors and Reasoning Operators}

Overall, Mobius achieves these two innate talents by decoupling the knowledge storage module, the FFN, from its layer-wise binding, and constructing a globally shared knowledge-vector database. First, this grants all reasoning operators, the Self-Attention, read access to the entire body of knowledge. Further, the Self-Attention gains the opportunity to access all knowledge within significantly fewer layers, and performs adaptive multi-round iteration through recurrent latent reasoning via Hidden States, ultimately emitting refined, high-information-density tokens in a single burst.

In constructing the shared knowledge repository, we opted for a straightforward horizontal concatenation, as the FFN is conventionally regarded as a Key-Value Knowledge-Vector Pool, and this concatenation approach preserves the correspondence relationships among these knowledge vectors. Although such simple concatenation yields performance benefits, excessive parameter activation degrades the model's training and inference efficiency. Therefore, at larger parameter scales, we employ a block-wise partitioning technique analogous to MoE~\citep{shazeer2017moe} to partition the FFN, with sparse activation during forward propagation.
\section{How about Mobius' performance?} \label{sec:experiments}

To validate the performance of Mobius, we conducted both training-from-scratch (TFS) and continual pre-training (CPT) experiments. For the TFS experiments, we trained 7B-A1B MoE models — one with Mobius and one with Transformer — on 1TB tokens, and compared their MMLU scores. For the CPT experiments, we used Qwen3.5-35B-A3B as the starting checkpoint, continued pre-training on 1TB tokens, and subsequently performed supervised fine-tuning (SFT) and reinforcement learning (RL).

\subsection{Mobius delivers markedly better data efficiency than Transformer.}

As shown in Figure \ref{fig:scratch-mmlu}, Mobius achieves significantly higher MMLU scores than Transformer across all stages of training. Moreover, when using Transformer's score at 1TB tokens as the baseline, Mobius attains the same score with only 0.626× data — that is, Mobius exhibits 1.6× data efficiency of Transformer.

The mechanism by which Mobius achieves higher data efficiency remains to be fully explored. Nevertheless, we hypothesize that under the Transformer architecture, parameters across layers exhibit considerable redundancy—for instance, the same piece of knowledge may be redundantly stored in multiple layers. Upon switching to the knowledge-reasoning-decoupled Mobius architecture, the model can attain superior compression rates, thereby enabling it to acquire sufficient knowledge with substantially less training data.

\subsection{Mobius matches Transformer-level reasoning with far greater inference efficiency.}

Given the prohibitive cost of pre-training from scratch, to validate Mobius's capabilities at scale, we chose to continue training from an existing open-source base model. As shown in Table \ref{tab:cpt-downstream}, starting from Qwen3.5 and switching to the Mobius architecture for continued pre-training not only preserves but actually enhances the model's overall capabilities.

More notably, switching to Mobius not only preserves fundamental reasoning capabilities but also improves end-to-end inference efficiency. As shown in Figure \ref{fig:cpt-efficiency}, aggregated across multiple evaluation benchmarks, Mobius achieves substantially higher request throughput than Transformer. We also examined the inference length of both models, visualized in Figure \ref{fig:cpt-cot}, which reveals that the improvement in Mobius's end-to-end inference efficiency primarily stems from shorter CoT lengths. When presented with identical problems, Mobius resolves them with markedly shorter reasoning chains, whereas Transformer requires longer CoT for deliberation. 

Although the precise mechanism behind Mobius's shorter CoT reasoning remains not fully established, we hypothesize that this may be attributed to Mobius's native latent reasoning characteristic, which internalizes the deliberation process within the model and achieves more efficient reasoning through continuously differentiable optimization.

However, whether Mobius merely hacks the problem or genuinely provides a more efficient mode of reasoning remains to be determined. By comparing multiple cases, we find that the shortened CoT in Mobius stems from the latter. The case in Table \ref{tab:linear-algebra-cot-comparison} is from MMLU-Pro Mathematics and tests two basic linear-algebra statements. The first statement is false because two vectors in \(\mathbb{R}^2\) need not be linearly independent. The second statement is true because a linearly independent set that spans \(V\) is a basis of \(V\), so its \(k\) vectors imply \(\dim(V)=k\). The correct answer is therefore Option C: False, True. 

\section{Mobius' relationship with mainstream research}

\begin{figure}[h]
    \centering
    \includegraphics[width=\textwidth]{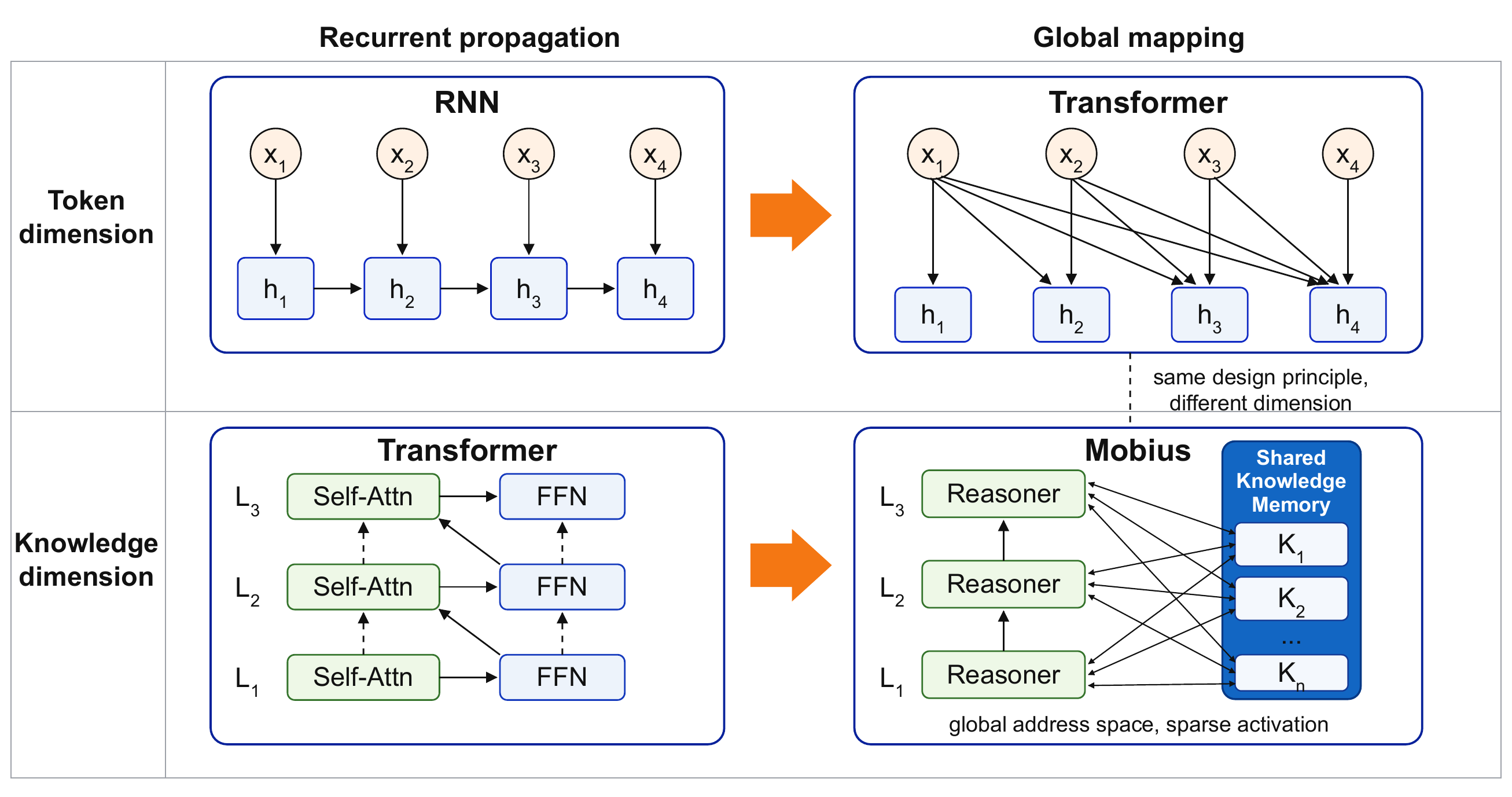} 
    \caption{The comparison between RNN, Transformer, and Mobius.}
    \label{fig:toy-genralization}
\end{figure}

\subsection{Latent Reasoning}

Chain-of-thought (CoT) externalizes reasoning into intermediate tokens~\citep{Wei0SBIXCLZ22}, but incurs sequential decoding overhead. Latent reasoning instead performs computation in continuous states, exploring three directions: continuous thought, looped computation, and parallel refinement.

\subsubsection{Continuous Thought}

Continuous-thought approaches replace discrete reasoning tokens with differentiable representations passed between reasoning steps. COCONUT directly feeds the final hidden state back as the next input embedding, allowing intermediate computation to proceed without decoding each state into language~\citep{hao2024coconut}. CODI compresses explicit CoT supervision into continuous states through self-distillation~\citep{shen2025codi}, while SoftCoT generates instance-specific soft thoughts with a lightweight assistant and projects them into the representation space of a target language model~\citep{xu2025softcot}. These approaches demonstrate that continuous states can preserve useful reasoning information while shortening or eliminating explicit rationales.

These approaches expose a small set of latent states that subsequent tokens can attend to and reuse as shared context. Mobius further makes this pattern a native architectural capability: its recurrent states are refined against the shared Memory and jointly support multiple future tokens, yielding higher-density reasoning without verbose traces.

\subsubsection{Looped Language Models}

Looped language models increase effective depth by repeatedly applying shared computation blocks. Universal Transformers introduced recurrent refinement across depth with parameter sharing~\citep{DehghaniGVUK19}, while subsequent analyses established the computational expressivity of Looped Transformers~\citep{loopedtransformer}. More recent work connects effective depth directly to reasoning: looped models can emulate multi-step latent computation and approach much deeper non-looped models on reasoning tasks~\citep{saunshi2025latent}, while recurrent-depth pre-training enables test-time compute scaling by increasing the number of latent iterations without generating additional reasoning tokens~\citep{geiping2025recurrent}.

Mobius also scales computation through latent recurrence, but performs each update over only a few layers while retaining access to the full shared Memory. This higher update frequency enables more iterative refinement and concentrates reasoning into higher-information-density latent states.

\subsubsection{Additional Latent Computation Steps}
Another line of work increases reasoning capacity by introducing extra latent computation steps before generation. Pause-token methods add learned placeholders to defer prediction and create additional computation space~\citep{goyal2024pause}. Hidden Decoding expands each token into multiple latent streams within a single forward pass, using intermediate key--value states as reusable computation traces~\citep{liu2026hidden}. Diffusion language models achieve similar effects through iterative refinement of continuous or masked representations~\citep{li2022diffusionlm,nie2025llada}. These methods shift computation from explicit CoT traces to latent trajectories.

Mobius extends this idea through recurrent latent refinement without explicit placeholders or diffusion steps. Its Reasoners repeatedly retrieve from a shared knowledge-vector Memory, refine high-density latent states, and decode multiple tokens in parallel.

\subsection{Efficient Reasoning}

Historically, the predominant paradigm for language model inference relied on autoregressive single-token generation from a single strong model, augmented with long chains of thought. To improve end-to-end inference efficiency, two approaches have gained increasing traction: Speculative Decoding, which enhances inference parallelism, and Concise CoT, which reduces the sequential length of reasoning chains.

\subsubsection{Speculative Decoding}

Speculative decoding accelerates autoregressive generation by using a cheap drafter to propose a block, or a tree, of future tokens and verifying these candidates in parallel with the target model \citep {stern2018blockwise,leviathan2023fast,chen2023speculative,miao2024specinfer}. With rejection-sampling correction, this verification preserves the target model's output distribution. While classical methods employ a lightweight external draft model, later work develops internal drafters based on auxiliary decoding heads or predicted hidden features \citep {cai2024medusa,li2024eagle,ankner2024hydra}. MTP is such an internal drafting mechanism rather than a decoding protocol itself: auxiliary heads sharing the target backbone predict multiple future tokens \citep {gloeckle2024better,liu2024deepseek,mehra2025multitoken}.

Whether employing a draft model or other strategies, Speculative Decoding traverses all knowledge only once before token prediction.  In contrast, Mobius performs multiple rounds of knowledge traversal and multi-token iteration internally, yielding hidden states of higher information density prior to final decoding, with native support for multi-token prediction.

\subsubsection{Concise Chain of Thought}
Long chain-of-thought (CoT) reasoning can improve performance on challenging tasks, but its autoregressive generation incurs substantial inference cost and latency. Recent work therefore seeks concise CoT through several complementary approaches: supervised methods learn from pruned or paired long--short trajectories, or internalize explicit rationales into implicit or dense representations \citep {kang2025c3ot,xia2025tokenskip,deng2024explicit,cheng2024compressed,deng2023implicit,yu2024distilling}; RL-based methods shape reasoning length through reward design, including controllable compression and length-aware objectives \citep {ma2025cotvalve,fatemi2025concise,song2025conciser,dai2025stable}. Although RL can elicit advanced behaviors such as self-correction, verification, and backtracking \citep {shao2024deepseekmath,deepseekai2025r1}, outcome-only rewards often favor longer trajectories because additional tokens enable further exploration and error correction~\citep{yang2025demystifying}. The central challenge is thus to reduce CoT verbosity without removing the deliberation needed for accurate and robust reasoning.

Conventional length regularization may shorten CoT by curtailing deliberation, potentially harming accuracy and robustness. In contrast, Mobius yields far more refined and shorter responses through several rounds of latent thought iteration.

\subsection{Architecture Design}


Mobius redesigns the internal information flow of Transformer architectures by decoupling knowledge storage from reasoning computation. 
In this section, we discuss its relationship with two major architectural directions: attention mechanisms and residual connections.

\subsubsection{Attention Mechanism}

Self-attention is the core component of Transformer architectures, enabling models to capture global dependencies through pairwise interactions among tokens~\citep{attentionisallyouneed}. However, the quadratic complexity of full attention has become a major bottleneck for scaling Transformer models to long-context scenarios. To address this issue, extensive studies have explored more efficient attention mechanisms. Sparse attention reduces unnecessary token interactions by restricting attention patterns, while linear attention reformulates the attention computation to reduce complexity through alternative kernelization or factorization strategies. More recently, state-space models and gated state-space variants, such as Mamba and Gated Delta Networks, replace explicit attention computation with recurrent state updates, providing efficient alternatives for long-sequence modeling~\citep{mamba,gdn}. These approaches mainly improve efficiency by modifying the computation form or reducing the cost of information interaction.

Rather than reducing attention complexity directly, we aim to increase the information density and value extracted per attention operation. Specifically, by decoupling knowledge from reasoning and constructing a globally shared knowledge repository, Mobius incur higher retrieval costs but provide attention with a more flexible selection space and higher-quality inputs, thereby completing reasoning with a more compact computation trajectory.

\subsubsection{Residual Connection}

Residual connections are fundamental components for scaling deep neural networks, enabling stable optimization and effective information propagation through shortcut pathways~\citep{he2016resnet}. Subsequent studies have explored more flexible connection patterns to improve information transmission across layers. Highway Networks introduce learnable gates to adaptively control information flow~\citep{srivastava2015highway}, while DenseNet increases connection density by allowing each layer to directly access previous representations~\citep{huang2017densely}. More recently, Hyper-Connections and Attention Residuals further investigate how to enhance residual pathways by increasing connection capacity or dynamically aggregating previous-layer representations~\citep{zhu2025hyperconnections,kimi2026attnres}. Despite these improvements, existing residual enhancement methods mainly focus on strengthening forward information propagation, where information flows from earlier layers to later layers.

Mobius extends residual design from forward information propagation toward bidirectional knowledge access, instead of explicitly expanding residual pathways. Since all reasoning stages can access the same knowledge repository, deeper Reasoners can retrieve knowledge beyond their local layer hierarchy, enabling more flexible information transmission while maintaining an efficient computation structure.


\section{Mobius' potential on several highlight topics}

\subsection{Self-Evolving}


Although models and agents have grown increasingly capable, how to enable them to continuously absorb new knowledge and skills from external sources remains an open question \cite{kirkpatrick2017overcoming,parisi2019continuallearning,WestonCB14,SantoroBBWL16}. 

Our position is that, \textbf{the current Transformer architecture does not satisfy the prerequisites for self-evolution.} Because knowledge and reasoning are tightly coupled in Transformer, such models can only acquire new skills through end-to-end training, which inevitably alters both knowledge and reasoning capabilities simultaneously, thereby causing catastrophic forgetting of previously learned skills. An architecture suitable for self-evolution should exhibit a certain degree of knowledge-reasoning decoupling, as this would support unbounded and efficient expansion of knowledge storage, and enable the model's reasoning capabilities to generalize across multiple domains. The Mobius architecture we propose possesses preliminary knowledge-reasoning separation characteristics, demonstrating greater potential for continual learning compared to Transformer. 

Nevertheless, whether Mobius can perform better in real-world self-evolution scenarios still requires the joint design of foundation models, agentic systems, and environmental feedback, and remains to be validated in future work.

\subsection{World Model}


World models are emerging as the next frontier beyond language models, yet how to enable models to perform understanding and reasoning over continuous-space inputs remains an open question \cite{ha2018worldmodels,hafner2019planet,hafner2023dreamer,schrittwieser2020muzero,BrownMRSKDNSSAA20,compute-optimal-llm}. 

Our position is that, \textbf{the current Transformer architecture does not satisfy the prerequisites for world models.} Models built primarily upon Transformer have been demonstrated to model discrete modalities (such as language) at the terabyte parameter scale, yet perfect modeling of continuous space with Transformer would likely require parameter scales on the order of petabytes. The inference cost of terabyte-scale language models already strains existing hardware to its limits; the inference cost of a petabyte-scale world model would be astronomical. The Mobius architecture we propose natively supports latent reasoning, offering a stronger prior for continuous-space modeling and demonstrating greater potential than Transformer. 

Nevertheless, the latent reasoning prior introduced by Mobius may prove far from sufficient; how to better reengineer the attention mechanism may be of even greater criticality.

\subsection{Scientific Discovery}



Contemporary AI has achieved remarkable proficiency in code and mathematics, and recent scientific foundation models excel at procedural problem-solving \cite{romera2024funsearch,boiko2023autonomous}. Nevertheless, enabling models to formulate disruptive scientific hypotheses and ideas remains an open challenge \cite{lu2024aiscientist,gottweis2026coscientist,wang2026naturebench}.

Our position is that, \textbf{the current paradigm of Transformer architecture combined with long chains of thought does not satisfy the prerequisites for scientific discovery.} Under the present paradigm, models excel at procedural problem-solving; however, scientific discovery demands not only procedural competence but also scientific intuition and the composition and generalization of knowledge. The Mobius architecture addresses this on two fronts: on one hand, it internalizes deliberation into a continuously differentiable latent space \cite{hao2024coconut,shen2025codi}, affording the model the opportunity to develop more powerful intuition; on the other hand, it flattens all knowledge within the model, enabling a greater volume of knowledge to interact simultaneously and substantially increasing the possibility of compositional generalization across knowledge domains. 

Nevertheless, whether Mobius can perform better in real-world scientific discovery scenarios likely requires the joint optimization of data, infrastructure, training algorithms, and optimization algorithms, and remains to be validated in future work.

\subsection{Hardware-Software Co-Design}


Models are being scaled to ever-larger parameter counts, demonstrating that greater scale does yield emergent intelligence, while simultaneously pushing the limits of physical hardware \cite{scalinglaws,compute-optimal-llm,BrownMRSKDNSSAA20,kwon2023pagedattention,rajbhandari2020zero}.

Our position is that, \textbf{the current combination of Transformer architecture and existing hardware systems does not satisfy the prerequisites for further scaling.} This is because the Transformer architecture commingles knowledge and reasoning within the same parameter set, necessitating the loading of nearly all parameters into GPU memory at deployment. The Mobius architecture we propose exhibits knowledge-reasoning separation, opening the possibility of stably retaining only reasoning-dedicated parameters in GPU memory while storing knowledge parameters predominantly on SSD, with high-priority knowledge retrieved and loaded into memory on demand.

In the long term, Mobius is more amenable to hardware-software co-design than Transformer. However, substantial effort remains before this vision can be realized.

\clearpage
\bibliographystyle{plain}
\bibliography{refs}

@misc{zou2026interns1pro,
  title        = {Intern-S1-Pro: Scientific Multimodal Foundation Model at Trillion Scale},
  author       = {Yicheng Zou and others},
  year         = {2026},
  eprint       = {2603.25040},
  archivePrefix= {arXiv},
  primaryClass = {cs.LG},
  url          = {https://arxiv.org/abs/2603.25040}
}

@misc{lu2024aiscientist,
  title         = {The AI Scientist: Towards Fully Automated Open-Ended Scientific Discovery},
  author        = {Lu, Chris and Lu, Cong and Lange, Robert Tjarko and Foerster, Jakob and Clune, Jeff and Ha, David},
  year          = {2024},
  eprint        = {2408.06292},
  archivePrefix = {arXiv},
  primaryClass  = {cs.AI},
  url           = {https://arxiv.org/abs/2408.06292}
}

@article{gottweis2026coscientist,
  title   = {Accelerating Scientific Discovery with Co-Scientist},
  author  = {Gottweis, Juraj and Weng, Wei-Hung and Daryin, Alexander and Tu, Tao and others},
  journal = {Nature},
  volume  = {655},
  number  = {8122},
  pages   = {487--496},
  year    = {2026},
  doi     = {10.1038/s41586-026-10644-y},
  url     = {https://doi.org/10.1038/s41586-026-10644-y}
}

@article{boiko2023autonomous,
  title   = {Autonomous Chemical Research with Large Language Models},
  author  = {Boiko, Daniil A. and MacKnight, Robert and Kline, Ben and Gomes, Gabe},
  journal = {Nature},
  volume  = {624},
  pages   = {570--578},
  year    = {2023},
  doi     = {10.1038/s41586-023-06792-0},
  url     = {https://doi.org/10.1038/s41586-023-06792-0}
}

@misc{wang2026naturebench,
  title         = {NatureBench: Can Coding Agents Match the Published SOTA of Nature-Family Papers?},
  author        = {Wang, Yuru and Cheng, Lejun and Zuo, Yuxin and Zeng, Sihang
                   and He, Bingxiang and Jiang, Che and Yang, Junlin
                   and Wang, Yuchong and Zhao, Kaikai and Huang, Weifeng
                   and Tian, Kai and Yuan, Zhenzhao and Zhong, Jincheng
                   and Wang, Weizhi and Ding, Ning and Zhou, Bowen
                   and Zhang, Kaiyan},
  year          = {2026},
  eprint        = {2606.24530},
  archivePrefix = {arXiv},
  primaryClass  = {cs.CL},
  doi           = {10.48550/arXiv.2606.24530},
  url           = {https://arxiv.org/abs/2606.24530}
}

@inproceedings{attentionisallyouneed,
 author = {Vaswani, Ashish and Shazeer, Noam and Parmar, Niki and Uszkoreit, Jakob and Jones, Llion and Gomez, Aidan N and Kaiser, \L ukasz and Polosukhin, Illia},
 booktitle = {Advances in Neural Information Processing Systems},
 editor = {I. Guyon and U. Von Luxburg and S. Bengio and H. Wallach and R. Fergus and S. Vishwanathan and R. Garnett},
 pages = {},
 publisher = {Curran Associates, Inc.},
 title = {Attention is All you Need},
 url = {https://proceedings.neurips.cc/paper_files/paper/2017/file/3f5ee243547dee91fbd053c1c4a845aa-Paper.pdf},
 volume = {30},
 year = {2017}
}

@misc{mhc,
      title={mHC: Manifold-Constrained Hyper-Connections}, 
      author={Zhenda Xie and Yixuan Wei and Huanqi Cao and Chenggang Zhao and Chengqi Deng and Jiashi Li and Damai Dai and Huazuo Gao and Jiang Chang and Kuai Yu and Liang Zhao and Shangyan Zhou and Zhean Xu and Zhengyan Zhang and Wangding Zeng and Shengding Hu and Yuqing Wang and Jingyang Yuan and Lean Wang and Wenfeng Liang},
      year={2026},
      eprint={2512.24880},
      archivePrefix={arXiv},
      primaryClass={cs.CL},
      url={https://arxiv.org/abs/2512.24880}, 
}

@article{methmatical-framework-for-transforner-circuits,
   title={A Mathematical Framework for Transformer Circuits},
   author={Elhage, Nelson and Nanda, Neel and Olsson, Catherine and Henighan, Tom and Joseph, Nicholas and Mann, Ben and Askell, Amanda and Bai, Yuntao and Chen, Anna and Conerly, Tom and DasSarma, Nova and Drain, Dawn and Ganguli, Deep and Hatfield-Dodds, Zac and Hernandez, Danny and Jones, Andy and Kernion, Jackson and Lovitt, Liane and Ndousse, Kamal and Amodei, Dario and Brown, Tom and Clark, Jack and Kaplan, Jared and McCandlish, Sam and Olah, Chris},
   year={2021},
   journal={Transformer Circuits Thread},
   note={https://transformer-circuits.pub/2021/framework/index.html}
}

@inproceedings{rome,
 author = {Meng, Kevin and Bau, David and Andonian, Alex and Belinkov, Yonatan},
 booktitle = {Advances in Neural Information Processing Systems},
 doi = {10.52202/068431-1262},
 editor = {S. Koyejo and S. Mohamed and A. Agarwal and D. Belgrave and K. Cho and A. Oh},
 pages = {17359--17372},
 publisher = {Curran Associates, Inc.},
 title = {Locating and Editing Factual Associations in GPT},
 url = {https://proceedings.neurips.cc/paper_files/paper/2022/file/6f1d43d5a82a37e89b0665b33bf3a182-Paper-Conference.pdf},
 volume = {35},
 year = {2022}
}

@inproceedings{key-value-memories,
    title = "Transformer Feed-Forward Layers Are Key-Value Memories",
    author = "Geva, Mor  and
      Schuster, Roei  and
      Berant, Jonathan  and
      Levy, Omer",
    editor = "Moens, Marie-Francine  and
      Huang, Xuanjing  and
      Specia, Lucia  and
      Yih, Scott Wen-tau",
    booktitle = "Proceedings of the 2021 Conference on Empirical Methods in Natural Language Processing",
    month = nov,
    year = "2021",
    address = "Online and Punta Cana, Dominican Republic",
    publisher = "Association for Computational Linguistics",
    url = "https://aclanthology.org/2021.emnlp-main.446/",
    doi = "10.18653/v1/2021.emnlp-main.446",
    pages = "5484--5495"
}

@misc{linearattention-tradeoff,
      title={Simple linear attention language models balance the recall-throughput tradeoff}, 
      author={Simran Arora and Sabri Eyuboglu and Michael Zhang and Aman Timalsina and Silas Alberti and Dylan Zinsley and James Zou and Atri Rudra and Christopher Ré},
      year={2025},
      eprint={2402.18668},
      archivePrefix={arXiv},
      primaryClass={cs.CL},
      url={https://arxiv.org/abs/2402.18668}, 
}

@misc{reasoningmodelseffectivethinking,
      title={Reasoning Models Can Be Effective Without Thinking}, 
      author={Wenjie Ma and Jingxuan He and Charlie Snell and Tyler Griggs and Sewon Min and Matei Zaharia},
      year={2025},
      eprint={2504.09858},
      archivePrefix={arXiv},
      primaryClass={cs.AI},
      url={https://arxiv.org/abs/2504.09858}, 
}

@inproceedings{
mamba,
title={Mamba: Linear-Time Sequence Modeling with Selective State Spaces},
author={Albert Gu and Tri Dao},
booktitle={First Conference on Language Modeling},
year={2024},
url={https://openreview.net/forum?id=tEYskw1VY2}
}

@inproceedings{simple-test-time-scaling,
    title = "s1: Simple test-time scaling",
    author = "Muennighoff, Niklas  and
      Yang, Zitong  and
      Shi, Weijia  and
      Li, Xiang Lisa  and
      Fei-Fei, Li  and
      Hajishirzi, Hannaneh  and
      Zettlemoyer, Luke  and
      Liang, Percy  and
      Cand{\`e}s, Emmanuel  and
      Hashimoto, Tatsunori",
    editor = "Christodoulopoulos, Christos  and
      Chakraborty, Tanmoy  and
      Rose, Carolyn  and
      Peng, Violet",
    booktitle = "Proceedings of the 2025 Conference on Empirical Methods in Natural Language Processing",
    month = nov,
    year = "2025",
    address = "Suzhou, China",
    publisher = "Association for Computational Linguistics",
    url = "https://aclanthology.org/2025.emnlp-main.1025/",
    doi = "10.18653/v1/2025.emnlp-main.1025",
    pages = "20275--20321",
    ISBN = "979-8-89176-332-6"
}

@inproceedings{
gdn,
title={Gated Delta Networks: Improving Mamba2 with Delta Rule},
author={Songlin Yang and Jan Kautz and Ali Hatamizadeh},
booktitle={The Thirteenth International Conference on Learning Representations},
year={2025},
url={https://openreview.net/forum?id=r8H7xhYPwz}
}

@misc{verifystepbystep,
      title={Let's Verify Step by Step}, 
      author={Hunter Lightman and Vineet Kosaraju and Yura Burda and Harri Edwards and Bowen Baker and Teddy Lee and Jan Leike and John Schulman and Ilya Sutskever and Karl Cobbe},
      year={2023},
      eprint={2305.20050},
      archivePrefix={arXiv},
      primaryClass={cs.LG},
      url={https://arxiv.org/abs/2305.20050}, 
}

@inproceedings{compute-optimal-llm,
    title = {Training Compute-Optimal Large Language Models},
    author = {Hoffmann, Jordan and Borgeaud, Sebastian and Mensch, Arthur and Buchatskaya, Elena and Cai, Trevor and Rutherford, Eliza and de Las Casas, Diego and Hendricks, Lisa Anne and Welbl, Johannes and Clark, Aidan and Hennigan, Tom and Noland, Eric and Millican, Katie and van den Driessche, George and Damoc, Bogdan and Guy, Aurelia and Osindero, Simon and Simonyan, Karen and Elsen, Erich and Vinyals, Oriol and Rae, Jack W. and Sifre, Laurent},
  booktitle = {Advances in Neural Information Processing Systems},
  volume = {35},
  pages = {30016--30030},
  year = {2022},
  publisher = {Curran Associates, Inc.}
}

@misc{scalinglaws,
      title={Scaling Laws for Neural Language Models}, 
      author={Jared Kaplan and Sam McCandlish and Tom Henighan and Tom B. Brown and Benjamin Chess and Rewon Child and Scott Gray and Alec Radford and Jeffrey Wu and Dario Amodei},
      year={2020},
      eprint={2001.08361},
      archivePrefix={arXiv},
      primaryClass={cs.LG},
      url={https://arxiv.org/abs/2001.08361}, 
}

@inproceedings{bert,
    title = "{BERT}: Pre-training of Deep Bidirectional Transformers for Language Understanding",
    author = "Devlin, Jacob  and
      Chang, Ming-Wei  and
      Lee, Kenton  and
      Toutanova, Kristina",
    editor = "Burstein, Jill  and
      Doran, Christy  and
      Solorio, Thamar",
    booktitle = "Proceedings of the 2019 Conference of the North {A}merican Chapter of the Association for Computational Linguistics: Human Language Technologies, Volume 1 (Long and Short Papers)",
    month = jun,
    year = "2019",
    address = "Minneapolis, Minnesota",
    publisher = "Association for Computational Linguistics",
    url = "https://aclanthology.org/N19-1423/",
    doi = "10.18653/v1/N19-1423",
    pages = "4171--4186"
}

@misc{vit,
      title={An Image is Worth 16x16 Words: Transformers for Image Recognition at Scale}, 
      author={Alexey Dosovitskiy and Lucas Beyer and Alexander Kolesnikov and Dirk Weissenborn and Xiaohua Zhai and Thomas Unterthiner and Mostafa Dehghani and Matthias Minderer and Georg Heigold and Sylvain Gelly and Jakob Uszkoreit and Neil Houlsby},
      year={2021},
      eprint={2010.11929},
      archivePrefix={arXiv},
      primaryClass={cs.CV},
      url={https://arxiv.org/abs/2010.11929}, 
}

@article{hao2024coconut,
  title={Training large language models to reason in a continuous latent space},
  author={Hao, Shibo and Sukhbaatar, Sainbayar and Su, DiJia and Li, Xian and Hu, Zhiting and Weston, Jason and Tian, Yuandong},
  journal={arXiv preprint arXiv:2412.06769},
  year={2024}
}

@article{liu2026hidden,
  title={Hidden Decoding at Scale: Latent Computation Scaling for Large Language Models},
  author={Liu, Aiwei and Shi, Cheng and Wu, Chuhan and Lei, Ci and Lu, Di and He, Donald and Zhang, Fan and Kong, Fanhao and Zhang, Feifei and Wang, Guan and others},
  journal={arXiv preprint arXiv:2607.08186},
  year={2026}
}

@inproceedings{xu2025softcot,
    title = "{S}oft{C}o{T}: Soft Chain-of-Thought for Efficient Reasoning with {LLM}s",
    author = "Xu, Yige  and
      Guo, Xu  and
      Zeng, Zhiwei  and
      Miao, Chunyan",
    editor = "Che, Wanxiang  and
      Nabende, Joyce  and
      Shutova, Ekaterina  and
      Pilehvar, Mohammad Taher",
    booktitle = "Proceedings of the 63rd Annual Meeting of the Association for Computational Linguistics (Volume 1: Long Papers)",
    month = jul,
    year = "2025",
    address = "Vienna, Austria",
    publisher = "Association for Computational Linguistics",
    url = "https://aclanthology.org/2025.acl-long.1137/",
    doi = "10.18653/v1/2025.acl-long.1137",
    pages = "23336--23351",
    ISBN = "979-8-89176-251-0"
}

@inproceedings{li2022diffusionlm,
 author = {Li, Xiang and Thickstun, John and Gulrajani, Ishaan and Liang, Percy S and Hashimoto, Tatsunori B},
 booktitle = {Advances in Neural Information Processing Systems},
 doi = {10.52202/068431-0313},
 editor = {S. Koyejo and S. Mohamed and A. Agarwal and D. Belgrave and K. Cho and A. Oh},
 pages = {4328--4343},
 publisher = {Curran Associates, Inc.},
 title = {Diffusion-LM Improves Controllable Text Generation},
 url = {https://proceedings.neurips.cc/paper_files/paper/2022/file/1be5bc25d50895ee656b8c2d9eb89d6a-Paper-Conference.pdf},
 volume = {35},
 year = {2022}
}

@inproceedings{nie2025llada,
 author = {Nie, Shen and Zhu, Fengqi and You, Zebin and Zhang, Xiaolu and Ou, Jingyang and Hu, Jun and Zhou, Jun and Lin, Yankai and Wen, Ji-Rong and LI, Chongxuan},
 booktitle = {Advances in Neural Information Processing Systems},
 editor = {D. Belgrave and C. Zhang and H. Lin and R. Pascanu and P. Koniusz and M. Ghassemi and N. Chen},
 pages = {50608--50646},
 publisher = {Curran Associates, Inc.},
 title = {Large Language Diffusion Models},
 url = {https://proceedings.neurips.cc/paper_files/paper/2025/file/48b383b24230e0e6e649d9c98dae4d8c-Paper-Conference.pdf},
 volume = {38},
 year = {2025}
}

@inproceedings{shen2025codi,
    title = "{CODI}: Compressing Chain-of-Thought into Continuous Space via Self-Distillation",
    author = "Shen, Zhenyi  and
      Yan, Hanqi  and
      Zhang, Linhai  and
      Hu, Zhanghao  and
      Du, Yali  and
      He, Yulan",
    editor = "Christodoulopoulos, Christos  and
      Chakraborty, Tanmoy  and
      Rose, Carolyn  and
      Peng, Violet",
    booktitle = "Proceedings of the 2025 Conference on Empirical Methods in Natural Language Processing",
    month = nov,
    year = "2025",
    address = "Suzhou, China",
    publisher = "Association for Computational Linguistics",
    url = "https://aclanthology.org/2025.emnlp-main.36/",
    doi = "10.18653/v1/2025.emnlp-main.36",
    pages = "677--693",
    ISBN = "979-8-89176-332-6"
}

@inproceedings{saunshi2025latent,
 author = {Saunshi, Nikunj and Dikkala, Nishanth and Li, Zhiyuan and Kumar, Sanjiv and J. Reddi, Sashank},
 booktitle = {International Conference on Learning Representations},
 editor = {Y. Yue and A. Garg and N. Peng and F. Sha and R. Yu},
 pages = {14855--14881},
 title = {Reasoning with Latent Thoughts: On the Power of Looped Transformers},
 url = {https://proceedings.iclr.cc/paper_files/paper/2025/file/2676109d49d1eb26d6bc584a8f556305-Paper-Conference.pdf},
 volume = {2025},
 year = {2025}
}

@inproceedings{geiping2025recurrent,
 author = {Geiping, Jonas and McLeish, Sean and Jain, Neel and Kirchenbauer, John and Singh, Siddharth and Bartoldson, Brian and Kailkhura, Bhavya and Bhatele, Abhinav and Goldstein, Tom},
 booktitle = {Advances in Neural Information Processing Systems},
 editor = {D. Belgrave and C. Zhang and H. Lin and R. Pascanu and P. Koniusz and M. Ghassemi and N. Chen},
 pages = {41340--41391},
 publisher = {Curran Associates, Inc.},
 title = {Scaling up Test-Time Compute with Latent Reasoning: A Recurrent Depth Approach},
 url = {https://proceedings.neurips.cc/paper_files/paper/2025/file/3b01972cf31e6fa0fe29e4b8b5c2a0a1-Paper-Conference.pdf},
 volume = {38},
 year = {2025}
}

@inproceedings{goyal2024pause,
  author = {Goyal, Sachin and Ji, Ziwei and Rawat, Ankit Singh and Menon, Aditya Krishna and Kumar, Sanjiv and Nagarajan, Vaishnavh},
 booktitle = {International Conference on Learning Representations},
 editor = {B. Kim and Y. Yue and S. Chaudhuri and K. Fragkiadaki and M. Khan and Y. Sun},
 pages = {27896--27923},
 title = {Think before you speak: Training Language Models With Pause Tokens},
 url = {https://proceedings.iclr.cc/paper_files/paper/2024/file/76917808731dae9e6d62c2a7a6afb542-Paper-Conference.pdf},
 volume = {2024},
 year = {2024}
}

@inproceedings{he2016identity,
  title={Identity mappings in deep residual networks},
  author={He, Kaiming and Zhang, Xiangyu and Ren, Shaoqing and Sun, Jian},
  booktitle={European conference on computer vision},
  pages={630--645},
  year={2016},
  organization={Springer}
}

@inproceedings{xiong2020layer,
  title={On layer normalization in the transformer architecture},
  author={Xiong, Ruibin and Yang, Yunchang and He, Di and Zheng, Kai and Zheng, Shuxin and Xing, Chen and Zhang, Huishuai and Lan, Yanyan and Wang, Liwei and Liu, Tieyan},
  booktitle={International conference on machine learning},
  pages={10524--10533},
  year={2020},
  organization={PMLR}
}

@article{stern2018blockwise,
  title={Blockwise parallel decoding for deep autoregressive models},
  author={Stern, Mitchell and Shazeer, Noam and Uszkoreit, Jakob},
  journal={Advances in Neural Information Processing Systems},
  volume={31},
  year={2018}
}

@article{ankner2024hydra,
  title={Hydra: Sequentially-dependent draft heads for medusa decoding},
  author={Ankner, Zachary and Parthasarathy, Rishab and Nrusimha, Aniruddha and Rinard, Christopher and Ragan-Kelley, Jonathan and Brandon, William},
  journal={arXiv preprint arXiv:2402.05109},
  year={2024}
}

@article{liu2024deepseek,
  title={Deepseek-v3 technical report},
  author={Liu, Aixin and Feng, Bei and Xue, Bing and Wang, Bingxuan and Wu, Bochao and Lu, Chengda and Zhao, Chenggang and Deng, Chengqi and Zhang, Chenyu and Ruan, Chong and others},
  journal={arXiv preprint arXiv:2412.19437},
  year={2024}
}

@inproceedings{leviathan2023fast,
  title={Fast inference from transformers via speculative decoding},
  author={Leviathan, Yaniv and Kalman, Matan and Matias, Yossi},
  booktitle={International Conference on Machine Learning},
  pages={19274--19286},
  year={2023},
  organization={PMLR}
}

@misc{chen2023speculative,
      title={Accelerating Large Language Model Decoding with Speculative Sampling}, 
      author={Charlie Chen and Sebastian Borgeaud and Geoffrey Irving and Jean-Baptiste Lespiau and Laurent Sifre and John Jumper},
      year={2023},
      eprint={2302.01318},
      archivePrefix={arXiv},
      primaryClass={cs.CL},
      url={https://arxiv.org/abs/2302.01318}, 
}

@inproceedings{miao2024specinfer,
  title={Specinfer: Accelerating large language model serving with tree-based speculative inference and verification},
  author={Miao, Xupeng and Oliaro, Gabriele and Zhang, Zhihao and Cheng, Xinhao and Wang, Zeyu and Zhang, Zhengxin and Wong, Rae Ying Yee and Zhu, Alan and Yang, Lijie and Shi, Xiaoxiang and others},
  booktitle={Proceedings of the 29th ACM International Conference on Architectural Support for Programming Languages and Operating Systems, Volume 3},
  pages={932--949},
  year={2024}
}

@misc{cai2024medusa,
      title={Medusa: Simple LLM Inference Acceleration Framework with Multiple Decoding Heads}, 
      author={Tianle Cai and Yuhong Li and Zhengyang Geng and Hongwu Peng and Jason D. Lee and Deming Chen and Tri Dao},
      year={2024},
      eprint={2401.10774},
      archivePrefix={arXiv},
      primaryClass={cs.LG},
      url={https://arxiv.org/abs/2401.10774}, 
}

@inproceedings{li2024eagle,
  title={EAGLE: speculative sampling requires rethinking feature uncertainty},
  author={Li, Yuhui and Wei, Fangyun and Zhang, Chao and Zhang, Hongyang},
  booktitle={Proceedings of the 41st International Conference on Machine Learning},
  pages={28935--28948},
  year={2024}
}

@inproceedings{gloeckle2024better,
  title={Better \& faster large language models via multi-token prediction},
  author={Gloeckle, Fabian and Idrissi, Badr Youbi and Rozi{\`e}re, Baptiste and Lopez-Paz, David and Synnaeve, Gabriel},
  booktitle={Proceedings of the 41st International Conference on Machine Learning},
  pages={15706--15734},
  year={2024}
}

@misc{mehra2025multitoken,
  title={On Multi-Token Prediction for Efficient LLM Inference},
  author={Somesh Mehra and Javier Alonso Garcia and Lukas Mauch},
  year={2025},
  eprint={2502.09419},
  archivePrefix={arXiv},
  primaryClass={cs.CL},
  url={https://arxiv.org/abs/2502.09419}
}

@inproceedings{kang2025c3ot,
  title={C3ot: Generating shorter chain-of-thought without compromising effectiveness},
  author={Kang, Yu and Sun, Xianghui and Chen, Liangyu and Zou, Wei},
  booktitle={Proceedings of the AAAI Conference on Artificial Intelligence},
  volume={39},
  number={23},
  pages={24312--24320},
  year={2025}
}

@inproceedings{xia2025tokenskip,
  title={Tokenskip: Controllable chain-of-thought compression in llms},
  author={Xia, Heming and Leong, Chak Tou and Wang, Wenjie and Li, Yongqi and Li, Wenjie},
  booktitle={Proceedings of the 2025 Conference on Empirical Methods in Natural Language Processing},
  pages={3351--3363},
  year={2025}
}

@article{deng2024explicit,
  title={From explicit cot to implicit cot: Learning to internalize cot step by step},
  author={Deng, Yuntian and Choi, Yejin and Shieber, Stuart},
  journal={arXiv preprint arXiv:2405.14838},
  year={2024}
}

@article{cheng2024compressed,
  title={Compressed chain of thought: Efficient reasoning through dense representations},
  author={Cheng, Jeffrey and Van Durme, Benjamin},
  journal={arXiv preprint arXiv:2412.13171},
  year={2024}
}

@misc{deng2023implicit,
  title={Implicit Chain of Thought Reasoning via Knowledge Distillation},
  author={Yuntian Deng and Kiran Prasad and Roland Fernandez and Paul Smolensky and Vishrav Chaudhary and Stuart Shieber},
  year={2023},
  eprint={2311.01460},
  archivePrefix={arXiv},
  primaryClass={cs.CL},
  url={https://arxiv.org/abs/2311.01460}
}

@misc{yu2024distilling,
  title={Distilling System 2 into System 1},
  author={Ping Yu and Jing Xu and Jason Weston and Ilia Kulikov},
  year={2024},
  eprint={2407.06023},
  archivePrefix={arXiv},
  primaryClass={cs.CL},
  url={https://arxiv.org/abs/2407.06023}
}

@misc{shao2024deepseekmath,
      title={DeepSeekMath: Pushing the Limits of Mathematical Reasoning in Open Language Models}, 
      author={Zhihong Shao and Peiyi Wang and Qihao Zhu and Runxin Xu and Junxiao Song and Xiao Bi and Haowei Zhang and Mingchuan Zhang and Y. K. Li and Y. Wu and Daya Guo},
      year={2024},
      eprint={2402.03300},
      archivePrefix={arXiv},
      primaryClass={cs.CL},
      url={https://arxiv.org/abs/2402.03300}, 
}

@article{deepseekai2025r1,
   title={DeepSeek-R1 incentivizes reasoning in LLMs through reinforcement learning},
   volume={645},
   ISSN={1476-4687},
   url={http://dx.doi.org/10.1038/s41586-025-09422-z},
   DOI={10.1038/s41586-025-09422-z},
   number={8081},
   journal={Nature},
   publisher={Springer Science and Business Media LLC},
   author={Guo, Daya and Yang, Dejian and Zhang, Haowei and Song, Junxiao and Wang, Peiyi and Zhu, Qihao and Xu, Runxin and Zhang, Ruoyu and Ma, Shirong and Bi, Xiao and Zhang, Xiaokang and Yu, Xingkai and Wu, Yu and Wu, Z. F. and Gou, Zhibin and Shao, Zhihong and Li, Zhuoshu and Gao, Ziyi and Liu, Aixin and Xue, Bing and Wang, Bingxuan and Wu, Bochao and Feng, Bei and Lu, Chengda and Zhao, Chenggang and Deng, Chengqi and Ruan, Chong and Dai, Damai and Chen, Deli and Ji, Dongjie and Li, Erhang and Lin, Fangyun and Dai, Fucong and Luo, Fuli and Hao, Guangbo and Chen, Guanting and Li, Guowei and Zhang, H. and Xu, Hanwei and Ding, Honghui and Gao, Huazuo and Qu, Hui and Li, Hui and Guo, Jianzhong and Li, Jiashi and Chen, Jingchang and Yuan, Jingyang and Tu, Jinhao and Qiu, Junjie and Li, Junlong and Cai, J. L. and Ni, Jiaqi and Liang, Jian and Chen, Jin and Dong, Kai and Hu, Kai and You, Kaichao and Gao, Kaige and Guan, Kang and Huang, Kexin and Yu, Kuai and Wang, Lean and Zhang, Lecong and Zhao, Liang and Wang, Litong and Zhang, Liyue and Xu, Lei and Xia, Leyi and Zhang, Mingchuan and Zhang, Minghua and Tang, Minghui and Zhou, Mingxu and Li, Meng and Wang, Miaojun and Li, Mingming and Tian, Ning and Huang, Panpan and Zhang, Peng and Wang, Qiancheng and Chen, Qinyu and Du, Qiushi and Ge, Ruiqi and Zhang, Ruisong and Pan, Ruizhe and Wang, Runji and Chen, R. J. and Jin, R. L. and Chen, Ruyi and Lu, Shanghao and Zhou, Shangyan and Chen, Shanhuang and Ye, Shengfeng and Wang, Shiyu and Yu, Shuiping and Zhou, Shunfeng and Pan, Shuting and Li, S. S. and Zhou, Shuang and Wu, Shaoqing and Yun, Tao and Pei, Tian and Sun, Tianyu and Wang, T. and Zeng, Wangding and Liu, Wen and Liang, Wenfeng and Gao, Wenjun and Yu, Wenqin and Zhang, Wentao and Xiao, W. L. and An, Wei and Liu, Xiaodong and Wang, Xiaohan and Chen, Xiaokang and Nie, Xiaotao and Cheng, Xin and Liu, Xin and Xie, Xin and Liu, Xingchao and Yang, Xinyu and Li, Xinyuan and Su, Xuecheng and Lin, Xuheng and Li, X. Q. and Jin, Xiangyue and Shen, Xiaojin and Chen, Xiaosha and Sun, Xiaowen and Wang, Xiaoxiang and Song, Xinnan and Zhou, Xinyi and Wang, Xianzu and Shan, Xinxia and Li, Y. K. and Wang, Y. Q. and Wei, Y. X. and Zhang, Yang and Xu, Yanhong and Li, Yao and Zhao, Yao and Sun, Yaofeng and Wang, Yaohui and Yu, Yi and Zhang, Yichao and Shi, Yifan and Xiong, Yiliang and He, Ying and Piao, Yishi and Wang, Yisong and Tan, Yixuan and Ma, Yiyang and Liu, Yiyuan and Guo, Yongqiang and Ou, Yuan and Wang, Yuduan and Gong, Yue and Zou, Yuheng and He, Yujia and Xiong, Yunfan and Luo, Yuxiang and You, Yuxiang and Liu, Yuxuan and Zhou, Yuyang and Zhu, Y. X. and Huang, Yanping and Li, Yaohui and Zheng, Yi and Zhu, Yuchen and Ma, Yunxian and Tang, Ying and Zha, Yukun and Yan, Yuting and Ren, Z. Z. and Ren, Zehui and Sha, Zhangli and Fu, Zhe and Xu, Zhean and Xie, Zhenda and Zhang, Zhengyan and Hao, Zhewen and Ma, Zhicheng and Yan, Zhigang and Wu, Zhiyu and Gu, Zihui and Zhu, Zijia and Liu, Zijun and Li, Zilin and Xie, Ziwei and Song, Ziyang and Pan, Zizheng and Huang, Zhen and Xu, Zhipeng and Zhang, Zhongyu and Zhang, Zhen},
   year={2025},
   month=Sept, pages={633–638} }

@inproceedings{yang2025demystifying,
  title={Demystifying Long Chain-of-Thought Reasoning},
  author={Yang, Shiming and Tong, Yuxuan and Niu, Xinyao and Neubig, Graham and Yue, Xiang},
  booktitle={International Conference on Machine Learning},
  pages={71177--71209},
  year={2025},
  organization={PMLR}
}

@misc{fatemi2025concise,
      title={Concise Reasoning via Reinforcement Learning}, 
      author={Mehdi Fatemi and Banafsheh Rafiee and Mingjie Tang and Kartik Talamadupula},
      year={2025},
      eprint={2504.05185},
      archivePrefix={arXiv},
      primaryClass={cs.CL},
      url={https://arxiv.org/abs/2504.05185}, 
}

@inproceedings{ma2025cotvalve,
  title={Cot-valve: Length-compressible chain-of-thought tuning},
  author={Ma, Xinyin and Wan, Guangnian and Yu, Runpeng and Fang, Gongfan and Wang, Xinchao},
  booktitle={Proceedings of the 63rd Annual Meeting of the Association for Computational Linguistics (Volume 1: Long Papers)},
  pages={6025--6035},
  year={2025}
}

@misc{song2025conciser,
      title={Walk Before You Run! Concise LLM Reasoning via Reinforcement Learning}, 
      author={Mingyang Song and Mao Zheng},
      year={2025},
      eprint={2505.21178},
      archivePrefix={arXiv},
      primaryClass={cs.CL},
      url={https://arxiv.org/abs/2505.21178}, 
}

@misc{dai2025stable,
      title={Stable Reinforcement Learning for Efficient Reasoning}, 
      author={Muzhi Dai and Shixuan Liu and Qingyi Si},
      year={2025},
      eprint={2505.18086},
      archivePrefix={arXiv},
      primaryClass={cs.AI},
      url={https://arxiv.org/abs/2505.18086}, 
}

@inproceedings{he2016resnet,
  author       = {Kaiming He and
                  Xiangyu Zhang and
                  Shaoqing Ren and
                  Jian Sun},
  title        = {Deep Residual Learning for Image Recognition},
  booktitle    = {Proceedings of the IEEE Conference on Computer Vision and Pattern Recognition},
  pages        = {770--778},
  year         = {2016}
}

@inproceedings{srivastava2015highway,
  author       = {Rupesh K. Srivastava and
                  Klaus Greff and
                  J{\"u}rgen Schmidhuber},
  title        = {Training Very Deep Networks},
  booktitle    = {Advances in Neural Information Processing Systems},
  pages        = {2377--2385},
  year         = {2015}
}

@inproceedings{huang2017densely,
  author       = {Gao Huang and
                  Zhuang Liu and
                  Laurens van der Maaten and
                  Kilian Q. Weinberger},
  title        = {Densely Connected Convolutional Networks},
  booktitle    = {Proceedings of the IEEE Conference on Computer Vision and Pattern Recognition},
  pages        = {4700--4708},
  year         = {2017}
}

@inproceedings{zhu2025hyperconnections,
  author       = {Defa Zhu and
                  Hongzhi Huang and
                  Zihao Huang and
                  Yutao Zeng and
                  Yunyao Mao and
                  Banggu Wu and
                  Qiyang Min and
                  Xun Zhou},
  title        = {Hyper-Connections},
  booktitle    = {International Conference on Learning Representations},
  year         = {2025}
}

@article{kimi2026attnres,
  author       = {{Kimi Team}},
  title        = {Attention Residuals},
  journal      = {arXiv preprint arXiv:2603.15031},
  year         = {2026}
}

@inproceedings{video_transformer,
  author = {Gedas Bertasius and
            Heng Wang and
            Lorenzo Torresani},
  title = {Is Space-Time Attention All You Need for Video Understanding?},
  booktitle = {ICML},
  year = {2021}
}

@inproceedings{lineartransformer,
  author = {Angelos Katharopoulos and
            Apoorv Vyas and
            Nikolaos Pappas and
            François Fleuret},
  title = {Transformers are RNNs: Fast Autoregressive Transformers with Linear Attention},
  booktitle = {ICML},
  pages = {5156--5165},
  year = {2020}
}

@inproceedings{performer,
  author = {Krzysztof Choromanski and
            Valerii Likhosherstov and
            Xingyou Song and
            Richard Davis and
            Kyle Sarlos and
            Adrian Weller},
  title = {Rethinking Attention with Performers},
  booktitle = {ICLR},
  year = {2021}
}

@inproceedings{highway,
  author = {Rupesh K. Srivastava and
            Klaus Greff and
            Jürgen Schmidhuber},
  title = {Highway Networks},
  booktitle = {ICML Deep Learning Workshop},
  year = {2015}
}

@inproceedings{densenet,
  author = {Gao Huang and
            Zhuang Liu and
            Laurens van der Maaten and
            Kilian Q. Weinberger},
  title = {Densely Connected Convolutional Networks},
  booktitle = {CVPR},
  pages = {4700--4708},
  year = {2017}
}

@inproceedings{loopedtransformer,
  author = {Angeliki Giannou and
            Shashank Rajput and
            Jy-yong Sohn and
            Kangwook Lee and
            Jason D. Lee and
            Dimitris Papailiopoulos},
  title = {Looped Transformers as Programmable Computers},
  booktitle = {ICML},
  pages = {11398--11442},
  year = {2023}
}

@inproceedings{shazeer2017moe,
  author = {Noam Shazeer and
            Azalia Mirhoseini and
            Krzysztof Maziarz and
            Andy Davis and
            Quoc Le and
            Geoffrey Hinton and
            Jeff Dean},
  title = {Sparsely-Gated Mixture-of-Experts Layers},
  booktitle = {ICLR},
  year = {2017}
}

@article{parisi2019continuallearning,
  title={Continual Lifelong Learning with Neural Networks: A Review},
  author={Parisi, German I. and Kemker, Ronald and Part, Jose L. and Kanan, Christopher and Wermter, Stefan},
  journal={Neural Networks},
  volume={113},
  pages={54--71},
  year={2019},
  publisher={Elsevier},
  doi={10.1016/j.neunet.2019.01.012}
}

@article{kirkpatrick2017overcoming,
  title={Overcoming catastrophic forgetting in neural networks},
  author={Kirkpatrick, James and Pascanu, Razvan and Rabinowitz, Neil and Veness, Joel and Desjardins, Guillaume and Rusu, Andrei A. and Milan, Kieran and Quan, John and Ramalho, Tiago and Grabska-Barwinska, Agnieszka and others},
  journal={Proceedings of the National Academy of Sciences},
  volume={114},
  number={13},
  pages={3521--3526},
  year={2017},
  publisher={National Academy of Sciences},
  doi={10.1073/pnas.1611835114}
}

@inproceedings{ha2018worldmodels,
  title={World Models},
  author={Ha, David and Schmidhuber, J{\"u}rgen},
  booktitle={Advances in Neural Information Processing Systems},
  volume={31},
  year={2018}
}

@inproceedings{hafner2019planet,
  title={Learning Latent Dynamics for Planning from Pixels},
  author={Hafner, Danijar and Lillicrap, Timothy and Fischer, Ian and Villegas, Ruben and Ha, David and Lee, Honglak and Davidson, James},
  booktitle={International Conference on Machine Learning},
  pages={2555--2565},
  year={2019},
  publisher={PMLR}
}

@article{hafner2023dreamer,
  title={Mastering Diverse Domains through World Models},
  author={Hafner, Danijar and Pasukonis, J{\"u}rgen and Ba, Jimmy and Lillicrap, Timothy},
  journal={arXiv preprint arXiv:2301.04104},
  year={2023}
}

@article{schrittwieser2020muzero,
  title={Mastering Atari, Go, Chess and Shogi by Planning with a Learned Model},
  author={Schrittwieser, Julian and Antonoglou, Ioannis and Hubert, Thomas and Simonyan, Karen and Sifre, Laurent and Schmitt, Simon and Guez, Arthur and Lockhart, Edward and Hassabis, Demis and Silver, David and others},
  journal={Nature},
  volume={588},
  pages={604--609},
  year={2020},
  publisher={Springer Nature},
  doi={10.1038/s41586-020-03051-4}
}

@article{romera2024funsearch,
  title={FunSearch: Making New Discoveries in Mathematical Sciences Using Large Language Models},
  author={Romera-Paredes, Bernardino and Barekatain, Mohammadamin and Novikov, Alexander and Balog, Matej and Kumar, M. Pawan and Dupont, Emilien and Ruiz, Francisco J. R. and Ellenberg, Jordan and others},
  journal={Nature},
  volume={625},
  pages={468--475},
  year={2024},
  publisher={Springer Nature},
  doi={10.1038/s41586-023-06924-6}
}

@inproceedings{kwon2023pagedattention,
  title={Efficient Memory Management for Large Language Model Serving with PagedAttention},
  author={Kwon, Woosuk and Li, Zhuohan and Zhuang, Siyuan and Sheng, Ying and Zheng, Lianmin and Yu, Cody Hao and Gonzalez, Joseph and Zhang, Hao and Stoica, Ion},
  booktitle={Proceedings of the 29th Symposium on Operating Systems Principles},
  pages={611--626},
  year={2023},
  publisher={ACM}
}

@inproceedings{rajbhandari2020zero,
  title={ZeRO: Memory Optimizations Toward Training Trillion Parameter Models},
  author={Rajbhandari, Samyam and Rasley, Jeff and Ruwase, Olatunji and He, Yuxiong},
  booktitle={SC20: International Conference for High Performance Computing, Networking, Storage and Analysis},
  pages={1--16},
  year={2020},
  publisher={IEEE}
}

@inproceedings{Wei0SBIXCLZ22,
  author       = {Jason Wei and
                  Xuezhi Wang and
                  Dale Schuurmans and
                  Maarten Bosma and
                  Brian Ichter and
                  Fei Xia and
                  Ed H. Chi and
                  Quoc V. Le and
                  Denny Zhou},
  title        = {Chain-of-Thought Prompting Elicits Reasoning in Large Language Models},
  booktitle    = {NeurIPS},
  year         = {2022}
}

@inproceedings{BrownMRSKDNSSAA20,
  author       = {Tom B. Brown and
                  Benjamin Mann and
                  Nick Ryder and
                  Melanie Subbiah and
                  Jared Kaplan and
                  Prafulla Dhariwal and
                  Arvind Neelakantan and
                  Pranav Shyam and
                  Girish Sastry and
                  Amanda Askell and
                  Sandhini Agarwal and
                  Ariel Herbert{-}Voss and
                  Gretchen Krueger and
                  Tom Henighan and
                  Rewon Child and
                  Aditya Ramesh and
                  Daniel M. Ziegler and
                  Jeffrey Wu and
                  Clemens Winter and
                  Christopher Hesse and
                  Mark Chen and
                  Eric Sigler and
                  Mateusz Litwin and
                  Scott Gray and
                  Benjamin Chess and
                  Jack Clark and
                  Christopher Berner and
                  Sam McCandlish and
                  Alec Radford and
                  Ilya Sutskever and
                  Dario Amodei},
  title        = {Language Models are Few-Shot Learners},
  booktitle    = {NeurIPS},
  year         = {2020}
}

@inproceedings{DehghaniGVUK19,
  author       = {Mostafa Dehghani and
                  Stephan Gouws and
                  Oriol Vinyals and
                  Jakob Uszkoreit and
                  Lukasz Kaiser},
  title        = {Universal Transformers},
  booktitle    = {{ICLR}},
  publisher    = {OpenReview.net},
  year         = {2019}
}

@article{SantoroBBWL16,
  author       = {Adam Santoro and
                  Sergey Bartunov and
                  Matthew M. Botvinick and
                  Daan Wierstra and
                  Timothy P. Lillicrap},
  title        = {One-shot Learning with Memory-Augmented Neural Networks},
  journal      = {CoRR},
  volume       = {abs/1605.06065},
  year         = {2016}
}

@inproceedings{WestonCB14,
  author       = {Jason Weston and
                  Sumit Chopra and
                  Antoine Bordes},
  title        = {Memory Networks},
  booktitle    = {{ICLR}},
  year         = {2015}
}

@misc{AllenZhu2024Phy,
    author = {{Allen-Zhu}, Zeyuan},
    title = {{ICML 2024 Tutorial: Physics of Language Models}},
    year = {2024},
    month = {July},
    note = {Project page: \url{https://physics.allen-zhu.com/}}
}


\clearpage
\appendix
\section{Author List} \label{app:authors}

\subsection{Core Contributors (Sorted by Contribution)}

\noindent \textbf{Algorithm Design:} 

\begin{itemize}[leftmargin=2em, itemsep=0.5em, topsep=0pt]
    \item \textbf{Architecture:} Ermo Hua$^{*}$, Xiangyu Hong$^{*}$, Che Jiang$^{*}$, Baiting Wu, Cheng Liang, Youbang Sun, Biqing Qi, Qipeng Guo
    \item \textbf{Training:} Baiting Wu$^{*}$, Chengqi Lv$^{*}$, Ermo Hua$^{*}$, Xiangyu Hong$^{*}$, Weida Wang, Ning Ding, Wenwei Zhang
\end{itemize}

\noindent \textbf{Infrastructure:}

\begin{itemize}[leftmargin=2em, itemsep=0.5em, topsep=0pt]
    \item \textbf{Training:} Hanjing Wang$^{*}$, Xiangyu Hong$^{*}$, Yicheng Gu$^{*}$, Ermo Hua$^{*}$, Shan Yu, Haozheng Hou, Jianmin Qian, Jie Hou, Zhongbo Tian, Hui Wang
    \item \textbf{Inference:} Qian Yao$^{*}$, Baiting Wu$^{*}$, Ermo Hua$^{*}$, Jifeng Ding, Ningsheng Ma, Han Lv, Minxi Jin, Zhongbo Tian, Hui Wang
\end{itemize}

\noindent \textbf{Project Leader:} Ermo Hua

\noindent \textbf{Project Advisors:} Qi Zhang, Kai Chen, Dahua Lin, Bowen Zhou

\subsection{Full List (Sorted by Character)}

Kai Chen, Jifeng Ding, Ning Ding, Jiaye Ge, Lixin Gu, Yicheng Gu, Qipeng Guo, Ermo Hua, Haian Huang, Haozheng Hou, Jie Hou, Xiangyu Hong, Che Jiang, Minxi Jin, Cheng Liang, Dahua Lin, Dawei Liu, Kuikun Liu, Chengqi Lv, Haijun Lv, Han Lv, Ningsheng Ma, Biqing Qi, Jianmin Qian, Shiya Su, Youbang Sun, Huanze Tang, Zhongbo Tian, Hanjing Wang, Rui Wang, Ting Wang, Yi Wang, Baiting Wu, Jun Xu, Bowen Yang, Hui Wang, Weida Wang, Haochen Ye, Jiashuo Yu, Shan Yu, Xiaoyi Yu, Qirui Zeng, Qi Zhang, Ming Zhang, Wenwei Zhang, Bowen Zhou, Xinyu Zhou

\section{Expert Activation Patterns} \label{app:expert_activation}

\begin{figure}[h]
    \centering
    \includegraphics[width=\textwidth]{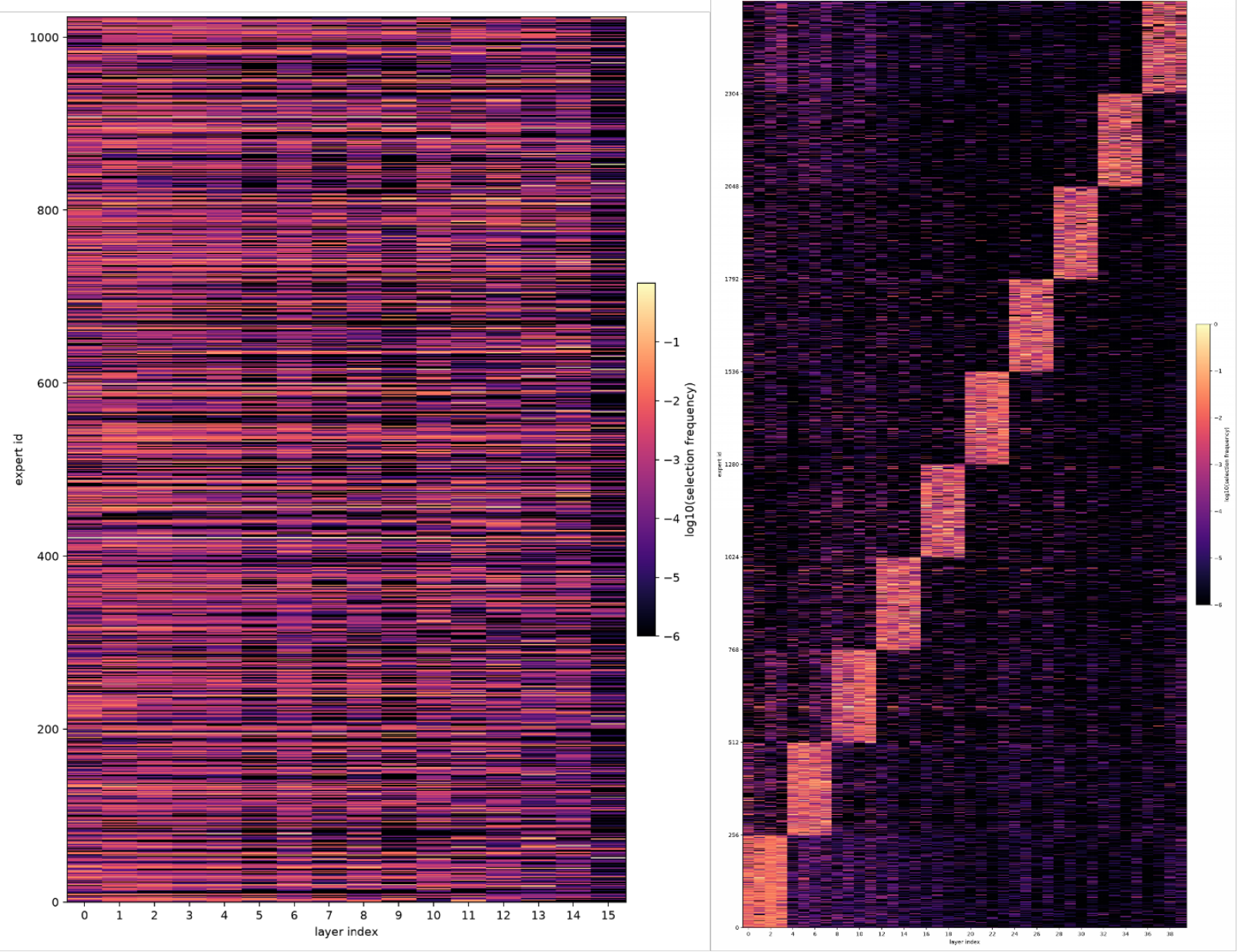} 
    \caption{Expert-Activation pattern across layers
    under two training recipes. The left panel shows Mobius-7B trained
    from scratch; the right panel shows Intern-S2-Mobius-35B obtained by
    architecture conversion and continual pre-training. Rows denote expert
    IDs, columns denote layer indices, and color represents the base-10
    logarithm of selection frequency. Expert IDs in the right panel are
    reordered to reveal the inherited routing structure.}
    \label{fig:expert-activation-patterns}
\end{figure}

We further compare expert-activation pattern under two training recipes
introduced in \ref{sec:experiments}. As shown in
\ref{fig:expert-activation-patterns}, the Mobius-7B model trained from scratch
(left) exhibits a comparatively uniform activation distribution across
reasoning layers. Intern-S2-Mobius-35B (right), continually pre-trained from
Qwen3.5-35B, activates experts over a wider
range while retaining a pronounced block-diagonal pattern. The latter suggests
that expert routing remains influenced by the layer-specific organization of
the source checkpoint.

These observations suggest that continual pre-training can broaden expert
utilization without fully removing the routing prior induced by architectural
conversion. Training Mobius from scratch may therefore permit more flexible
access to the shared expert pool; whether this translates into stronger model
capability requires controlled evaluation at comparable scale.

\section{Toy Task of Compositional Generalization} \label{app:expert_activation}

We select a Compositional Generalization task from Physics of LLM ~\citep{AllenZhu2024Phy}. On this task, we compared Transformer against our yet-to-be-released, newer Mobius architecture. As shown in Fig \ref{fig:toy-genralization}, Mobius achieved substantially better convergence efficiency and final scores. This compositional generalization task uses single-hop knowledge of 500 entities and two-hop knowledge of 400 entities as the training set, and two-hop knowledge of the remaining 100 entities as the test set. By comparing test-set scores, we examine whether the model can genuinely learn the connections between different knowledge pieces and achieve compositional generalization.

\begin{figure}[h]
    \centering
    \includegraphics[width=\textwidth]{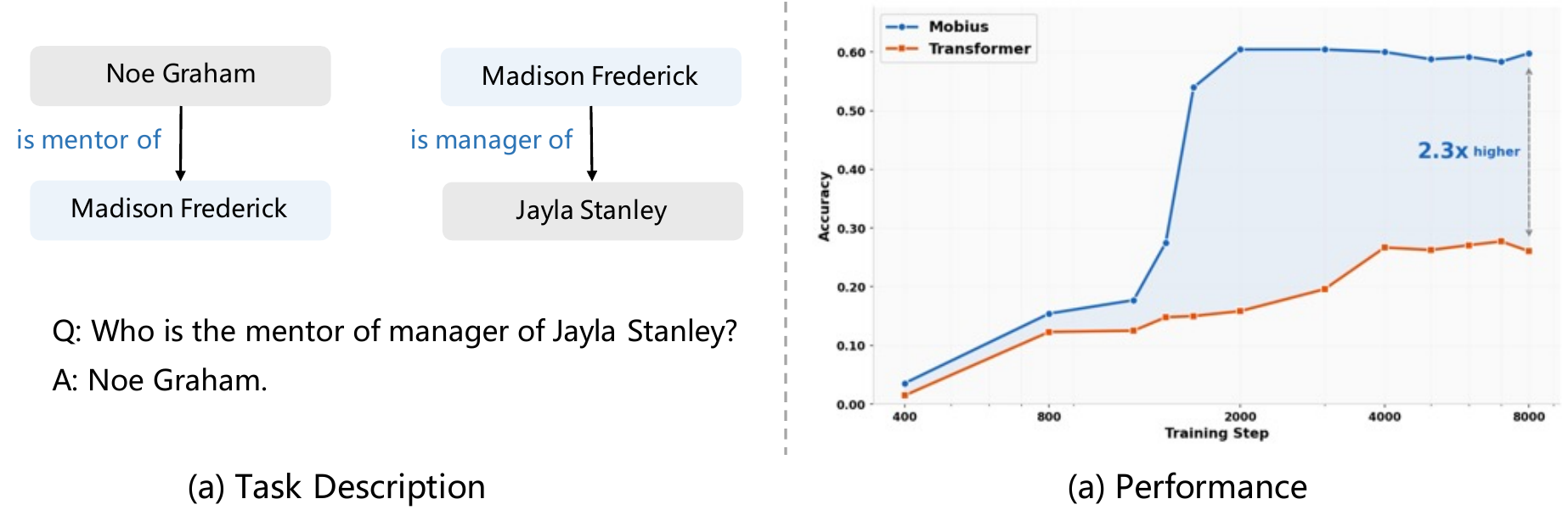} 
    \caption{The Compositional Generalization task selected from Physics of LLM. A future-version of Mobius(unreleased) enjoys faster convergence, higher final scores than Transformer. The train set contains 500-entity single-hop + 400-entity two-hop, while the test set contains 100-entity two-hop holdout. This task evaluates whether models learn cross-knowledge connections, not mere memorization.}
    \label{fig:toy-genralization}
\end{figure}

\section{Case Study: Chain-of-Thought Reasoning} \label{app:case_study}

We further present an example from the field of biology in \ref{tab:biology-cot-comparison}. In this case, Mobius also delivers a mush shorter reasoning trace than Transformer.

\begin{table*}[h]
    \centering
    \caption{Step-aligned comparison between Intern-Spin-35B and Qwen3.5-35B on a
    photosynthesis multiple-choice question. Both models select the correct
    answer (Option D). Token counts are computed using the Qwen3.5-35B tokenizer.}
    \label{tab:biology-cot-comparison}
    \small
    \setlength{\tabcolsep}{8pt}
    \renewcommand{\arraystretch}{1.15}

    \begin{adjustbox}{max width=\textwidth}
        \begin{tabular}{l B{5.5cm} C{5.5cm}}

        \SubTableTitle{Reasoning Trace Comparison}
        \toprule
        \textbf{Aligned Step} &
        \textbf{Intern-Spin-35B} &
        \textbf{Qwen3.5-35B} \\
        \midrule

        Task framing &
        \textbf{36 tokens.}

        ``The user is asking a question about the relationship between the light
        reactions and the Calvin cycle in photosynthesis.'' &
        \textbf{35 tokens.}

        ``The user wants me to answer \ldots{} \textbf{Question:} The light
        reactions \ldots{} supply the Calvin cycle with which of the following?''
        \\

        \addlinespace[0.25em]
        Core biological mechanism &
        \textbf{108 tokens.}

        ``Light-dependent reactions \ldots{} producing oxygen, ATP, and NADPH.''

        ``Calvin cycle \ldots{} uses ATP and NADPH \ldots{} to fix \(CO_2\)
        into sugars.'' &
        \textbf{276 tokens.}

        ``Light-dependent reactions \ldots{} Outputs: Oxygen (\(O_2\)), ATP,
        NADPH. \ldots{} Calvin Cycle \ldots{} Inputs: \(CO_2\), ATP, NADPH.''

        ``\ldots{} reduction requires ATP and NADPH \ldots{}''
        \\

        \addlinespace[0.25em]
        Option evaluation &
        \textbf{240 tokens.}

        ``D: ATP provides energy, NADPH provides high-energy electrons
        \ldots{}''

        ``H: ATP and NADPH are produced in light reactions, not broken down
        there \ldots{}'' &
        \textbf{543 tokens.}

        ``D \ldots{} Correct. The light reactions generate ATP (energy) and
        NADPH (reducing power/electrons). \ldots{}''

        ``H \ldots{} ATP and NADPH are produced in the light reactions and
        consumed \ldots{} in the Calvin cycle.''
        \\

        \addlinespace[0.25em]
        Repeated derivation &
        \textemdash &
        \textbf{205 tokens.}
        
        ``\textbf{Conclusion:} Option D accurately describes the relationship
        \ldots{} ATP and NADPH \ldots{} are used in the Calvin cycle to fix
        carbon.''
        
        ``\textbf{Step-by-step derivation:} \ldots{} Light Reactions produce
        ATP and NADPH \(\rightarrow\) Calvin Cycle consumes them. \ldots{}''

        \\

        \addlinespace[0.25em]
        Repeated option check &
        \textemdash &
        \textbf{143 tokens.}

        ``\textbf{Final check:} \ldots{} Option A: \(CO_2\) source is wrong.
        \ldots{} Option D: ATP/NADPH transfer is correct. \ldots{}''
        \\

        \addlinespace[0.25em]
        Reasoning closure &
        \textbf{13 tokens.}

        ``Conclusion: Option D is the correct answer.'' &
        \textbf{8 tokens.}

        ``The correct option is D.''
        \\

        \addlinespace[0.25em]
        Visible final answer &
        \textbf{75 tokens.}

        ``The light reactions \ldots{} stored in ATP and NADPH. These molecules
        are then used in the Calvin cycle \ldots{}''

        ``ANSWER: D'' &
        \textbf{439 tokens.}

        ``The process of photosynthesis is divided into two main stages
        \ldots{} ATP and NADPH \ldots{} are transported to the Calvin cycle.''

        ``\ldots{} D: This states ATP and NADPH are supplied. This is accurate.
        \ldots{}''

        ``ANSWER: D''
        \\

        \midrule
        \textbf{Total} &
        \textbf{472 tokens} &
        \textbf{1,649 tokens} \\

        \bottomrule
        \end{tabular}
    \end{adjustbox}

    \vspace{0.3em}
    \begin{minipage}{0.94\textwidth}
    \footnotesize
    \textit{Note:} The token counts are sequential marginal counts over each
    raw prediction; therefore, they sum exactly to the total output length.
    The displayed excerpts are representative original text from the
    corresponding reasoning segment. Ellipses indicate omitted original text.
    \end{minipage}
\end{table*}

\section{Layerwise Analysis of Latent Reasoning} \label{app:mtp_lens}

\begin{figure*}[t]
    \centering
    \includegraphics[width=\textwidth]{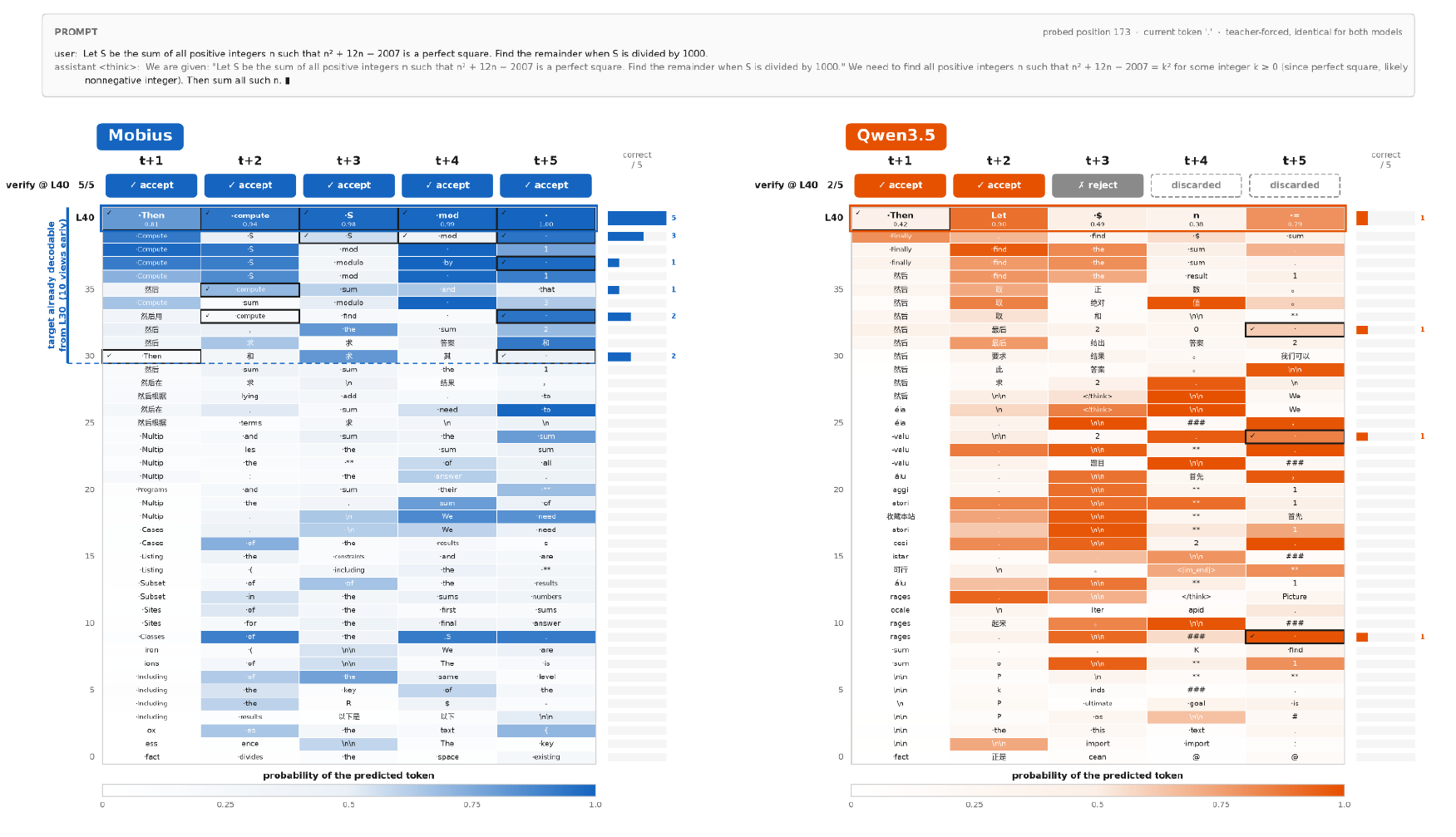}
    \caption{Layerwise prediction lens for Mobius and the Qwen3.5 baseline under an identical teacher-forced context. The $t+1$ column represents standard next-token prediction, while $t+2$ through $t+5$ are subsequent MTP predictions. Each cell shows the token predicted from a given hidden-state view; color intensity denotes its probability, and a black outline marks agreement with the target continuation. Mobius exhibits more target-aligned intermediate predictions and produces a five-token draft accepted in full, while the baseline produces only a two-token accepted prefix.}
    \label{fig:mtp_lens}
\end{figure*}

We use a layerwise prediction lens to examine how token predictions evolve during latent computation. Given the same teacher-forced context, we decode the hidden state at each layer for the standard next-token prediction ($t+1$) and four subsequent MTP positions ($t+2$ to $t+5$). As shown in \Cref{fig:mtp_lens}, Mobius exhibits more interpretable, target-aligned predictions in its intermediate layers than the baseline. Its layerwise predictions follow a more coherent semantic trajectory, suggesting that task-relevant information is concentrated into a compact internal representation rather than dispersed across competing token candidates. Further latent iterations progressively refine this representation and improve predictions at subsequent positions, ultimately yielding a fully accepted five-token draft. In contrast, the baseline exhibits less stable intermediate predictions, and its draft is rejected at the third predicted position, leaving only a two-token accepted prefix. This example suggests that Mobius may form more compact internal representations and refine them iteratively in latent space, potentially benefiting multi-token prediction.

\end{document}